\documentclass[runningheads]{llncs}
\usepackage{graphicx}
\usepackage{amsmath}
\usepackage{amssymb}
\usepackage{booktabs}
\usepackage{enumitem}
\usepackage{array}
\usepackage{wrapfig}
\usepackage{sidecap}
\sidecaptionvpos{figure}{c}
\usepackage{color}
\usepackage[toc,page]{appendix}
\usepackage{bbm}
\usepackage[noend]{algpseudocode} 
\usepackage{bbding}
\usepackage[normalem]{ulem}
\usepackage{bm}
\usepackage{amsfonts}       
\usepackage{multirow}
\usepackage{mathtools}
\usepackage{makecell}
\usepackage{rotating}
\usepackage{eqparbox}
\usepackage{longtable}
\usepackage{tabularx}
\usepackage{stfloats}
\usepackage{algorithm}
\usepackage{diagbox} 
\usepackage{esvect}
\usepackage{ragged2e} 
\usepackage{hyphenat}
\usepackage{balance}

\usepackage{tikz}
\usepackage{comment}
\usepackage{amsmath,amssymb} 
\usepackage{color}

\usepackage[accsupp]{axessibility}  

\usepackage[pagebackref,breaklinks,colorlinks]{hyperref}

\usepackage{sidecap}

\begin{document}
\pagestyle{headings}
\mainmatter

\title{SemAug: Semantically Meaningful Image Augmentations for Object Detection\\ Through Language Grounding} 

\titlerunning{SemAug: Semantically Meaningful Image Augmentation} 
\author{Morgan Heisler \and
Amin Banitalebi-Dehkordi \and
Yong Zhang}
\authorrunning{M. Heisler et al.}
\institute{Huawei Technologies Canada Co., Ltd., Burnaby, Canada \\
\email{\{morgan.lindsay.heisler, amin.banitalebi, yong.zhang3\}@huawei.com}}
\maketitle

\begin{abstract}
Data augmentation is an essential technique in improving the generalization of deep neural networks. The majority of existing image-domain augmentations either rely on geometric and structural transformations, or apply different kinds of photometric distortions. In this paper, we propose an effective technique for image augmentation by injecting contextually meaningful knowledge into the scenes. Our method of semantically meaningful image augmentation for object detection via language grounding, SemAug, starts by calculating semantically appropriate new objects that can be placed into relevant locations in the image (the \texttt{what} and \texttt{where} problems). Then it embeds these objects into their relevant target locations, thereby promoting diversity of object instance distribution. Our method allows for introducing new object instances and categories that may not even exist in the training set. Furthermore, it does not require the additional overhead of training a context network, so it can be easily added to existing architectures. Our comprehensive set of evaluations showed that the proposed method is very effective in improving the generalization, while the overhead is negligible. In particular, for a wide range of model architectures, our method achieved ~2-4\% and ~1-2\% mAP improvements for the task of object detection on the Pascal VOC and COCO datasets, respectively. Code is available as supplementary.

\keywords{Semantic Image Augmentation, Language Grounding}
\end{abstract}

\section{Introduction} \label{sec:introduction}

Training a deep neural network (DNN) relies on the availability of representative datasets which contain a sufficient number of labeled examples. Collecting relevant samples and labeling them is a time consuming and costly task. In practice, various techniques are employed to improve the network accuracy given the available training data. Of these techniques, methods of artificially expanding the size of the training dataset are of especial importance. For computer vision tasks, image augmentation is a technique that is used to artificially expand the size of a training dataset by generating modified versions of the training images. Almost all modern vision-DNNs involve some form of image augmentation in training \cite{shorten2019survey}. The importance of augmentation is even more pronounced for applications where training data is imbalanced (distribution of instances among categories is not uniform), when target categories are rare or uncommon in nature (e.g. detection of security threats), or when adding new object categories to datasets.

\begin{figure}[t!]
    \centering
    \includegraphics[width=0.98\linewidth]{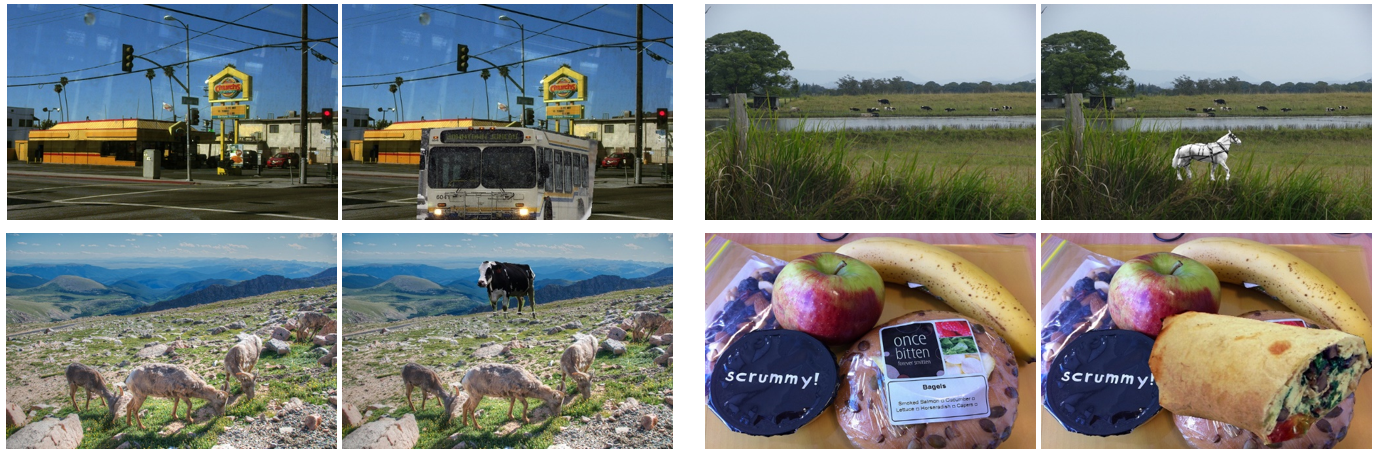}
    \caption{Examples of our method: originals (left) and semantically augmented (right).}
    \label{fig:nice_examples}
\end{figure}

Although traditional techniques such as flipping, rotating, cropping, and altering the colour space are helpful, they are generic and all-purpose in nature. When performing visual tasks such as object detection and semantic segmentation, a more object-aware method specifically created for these tasks could improve results. To address this, \cite{random},\cite{cutpastelearn} performed studies of placing objects randomly inside training images, and observed consistently better results for both object detection and semantic segmentation tasks.

Conversely, though randomly placing new object instances has an effect of generating more training samples and therefore reduces over-fitting, it is likely forcing the detector to fixate on the appearance of individual objects thereby becoming invariant to contextual information that humans find useful~\cite{context}. Intuitively, methods which preserve such context should further boost performance results. This intuition was confirmed by \cite{context}, \cite{context2018} that showed context-based object placement achieves higher generalization compared to random placement. However, training a context model adds a considerable overhead, making it less practical in real-world applications \cite{instaboost}. In addition, contextual associations in such methods are derived using data in the training dataset and therefore the potential for new associations is limited.

\begin{figure*}
    \centering
    \includegraphics[width=0.95\linewidth]{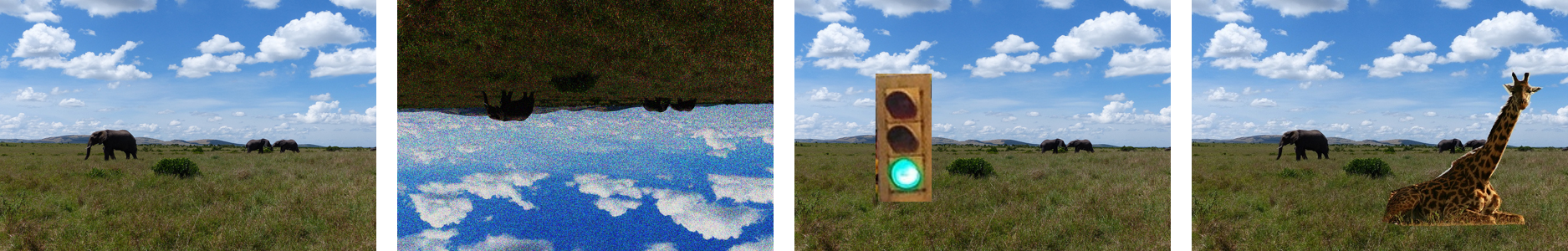}
    \caption{Various methods of augmentation: From left to right: the original image, traditional augmentations (flip, contrast/brightness adjustment, additive noise), random object placement, and SemAug (our method). A giraffe could reasonably be found in a field with elephants, whereas a traffic light has no contextual basis in this scene.}
    \label{fig:various_methods}
\end{figure*}

Contextual relationships have been an area of interest in the Natural Language Processing (NLP) world for quite some time \cite{allen1988natural,bert,hirschberg2015advances,gokhale-etal-2020-mutant}. In this domain, words can be represented as real-valued vectors allowing for quantitative latent space analysis. Various language models such as GloVe \cite{glove}, fasttext \cite{fasttext}, BERT~\cite{bert}, etc. have been trained on vast text corpora to encode the intricate relationships between words. In this paper, we present a simple and effective method for injecting contextual information using these semantic word vectors, without the overhead of training additional dataset-dependent context networks. By leveraging the semantic labels, our method can consider the context of a scene and augment appropriately as shown in Figure \ref{fig:nice_examples} through the injection of contextual knowledge. In brief, word vectors from pre-trained embeddings are used to compute the most similar objects which can then be placed in an image. A comparison to other techniques is shown in Figure \ref{fig:various_methods} where the original image consists of elephants in a field with sky above. Traditional techniques are able to globally modify the image to look different than the original, but do not add any information based on prior knowledge. Neither does the random object placement which placed a traffic light in the scene. The semantically augmented image has added a giraffe, which is contextually relevant, and added it to the scene in an appropriate location. This addition based on prior knowledge aids the network in discerning the relationships between objects in a dataset.

The main contributions of this paper are as follows:
\begin{itemize}[wide, labelindent=0.5pt]
    \item We present a new method of object-based contextually meaningful image augmentation for object detection. In particular, we propose a solution for the \texttt{what} and \texttt{where} problems for object instance placement. 
    Moreover, our method allows for the introduction of new object categories into a dataset while still considering context as it is not dataset-dependent.
    \item Our method considers context without the overhead of training additional context models, allowing for easy adoption to existing models and training pipelines.
    \item Through a comprehensive set of experiments, we show that our method provides consistent improvements on standard object detection benchmarks. We show our context-based object handling is indeed more meaningful than random placements, while it does not require training additional context models. 
\end{itemize}

\section{Related works} \label{sec:related_works}

Related to our work are image augmentation methods, in particular context-based augmentations. We provide a brief overview in this section.

\noindent\textbf{Traditional augmentations:} Include rotate, flip, resize, blur, added noise, color manipulations, and other geometric or photometric transformations. A typical preprocessing pipeline may include a combination of such augmentations.

\noindent\textbf{Combining image augmentations:} To address situations where traditional augmentations do not cover, several more advanced methods of mixing augmentations and their respective labels have been proposed in the recent years. Examples include: RandAugment~\cite{randaugment}/AutoAugment~\cite{autoaugment} (to identify suitable augmentations on each training iteration), AugMix~\cite{augmix} (mixing multiple random augmentations and enforcing a consistency loss), MixUp~\cite{mixup} (mixing training samples), CutOut \cite{cutout} (cutting out a random bounding box), CutMix \cite{cutmix} (cut a random box from an image and paste to another), DeepAugment~\cite{deepaugment} (adding perturbation on weights and activations), FenceMask~\cite{fencemask} (fence-shape CutOut), FMix~\cite{fmix} (applying binary masks from Fourier space), KeepAugment~\cite{keepaugment} (CutMix but not applying any augmentations on the pasted box), ClassMix~\cite{classmix} (combining segmentation masks), ComplexMix~\cite{complexmix} (advanced version of ClassMix). These augmentations have shown to improve on the traditional augmentations, however as mentioned in the introduction section, some form of object-level augmentation specifically created for the task of object detection rather than image classification may provide a larger performance boost.

\noindent\textbf{Object-level augmentations:} The previous methods do not consider any specific object-level augmentations but rather apply some transformations over the whole image, which may not be optimal for tasks such as object detection. Recently, object-aware methods such as Copy-Paste \cite{random} or Cut-Paste-Learn \cite{cutpastelearn} have gained traction (denoted by `random' in Figure \ref{fig:various_methods}). Though these methods do increase the number of object instances in a dataset, no prior contextual knowledge is used to determine whether pasted objects would naturally be found in the scene. This is the major disadvantage because the object-aware method may learn improper associations which would not appear in test images, leading to inevitable accuracy loss in object detection. 

Other than the random object pasting, some recent methods propose other approaches of object-level augmentation. InstaBoost \cite{instaboost} proposes to move an object within its neighborhood to create new training examples. Inpainting might be used to fill in the black pixels. PSIS \cite{psis} and COCP \cite{cocp} on the other hand, switch different instances of a same object category within two images. While effective, these methods provide sub-optimal augmentation as the object categories for each image do not change. Additionally, the constraints in place for COCP \cite{cocp} inhibit the number of synthetic images that can be created, especially for a smaller dataset. Our method is able to add new object categories to images, enabling stronger perturbations in the image domain, as well as add new object categories to the datasets.

\noindent\textbf{Contextual augmentations:}
To take context into consideration, Context-DA \cite{context,context2018} and \cite{volokitin2020efficiently} proposed to train a separate model that learns the context. The main disadvantages of using an additional context model are: additional networks require extra training overhead, and are highly dataset-dependent. Our method differs as it does not model the visual context of the images, but rather leverages the availability of pre-trained language embeddings to derive semantic context from images. This allows for the injection of new knowledge not necessarily already in the given dataset, less overhead than training and inferencing an additional context network, and improved flexibility as it can be readily used in any architecture or framework.
\section{Method} \label{sec:method}
In this section, we first provide a formulation of the problem. Then, we describe our method and provide insights on different aspects of our approach.

\subsection{Problem statement}
Let \(\mathbf{I} \in \mathbb{R}^{W\times H}\) denote a training image from the train set with width $W$ and height $H$ (for brevity we drop the channels dimension). The goal of SemAug is to generate a new training sample $\tilde{\mathbf{I}}$ by inserting one or more contextually relevant objects from an object bank $\Omega$ using semantic knowledge, $\pi$. This can be expressed by:
\begin{equation}
\label{eq:problem}
\tilde{\mathbf{I}} = f_\pi(\mathbf{I},\Omega).
\end{equation}
In this section, we present a method of language grounding as a way of extracting and matching high level semantic context $\pi$.
The augmented training sample set $\{\tilde{\mathbf{I}}\}$ is then used to train the model with its original loss function. Through this injection of semantic knowledge, we strengthen the network's ability to predict objects given context.

\subsection{Semantic augmentation}

\begin{figure*}[t!]
    \centering
    \includegraphics[width=1.0\linewidth]{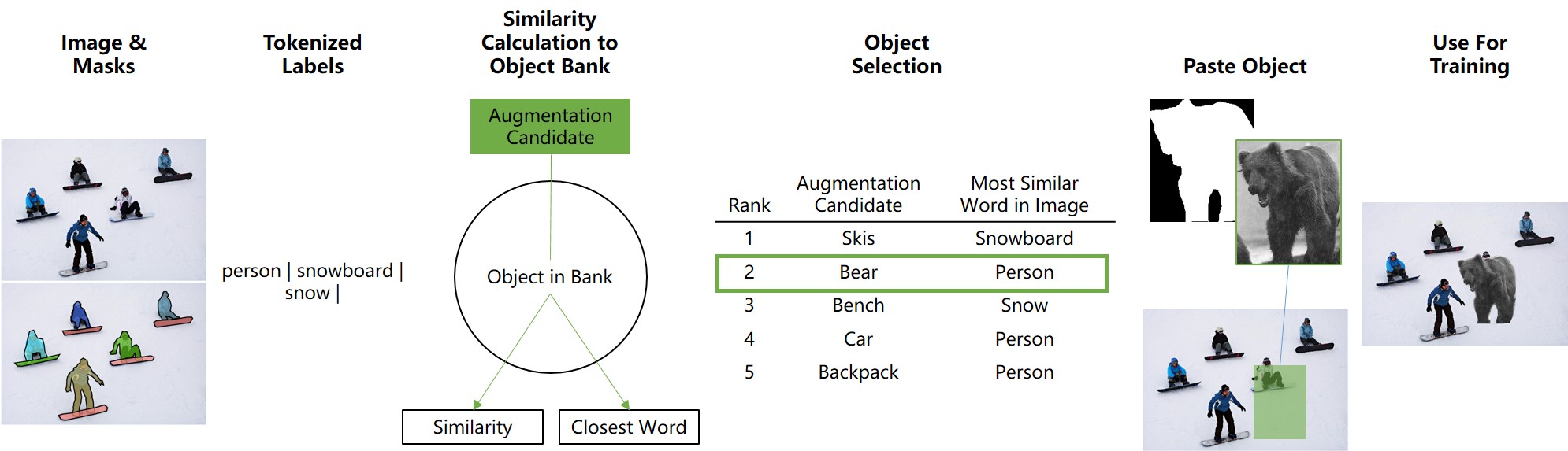}
    \caption{Illustration of our data augmentation approach. After an image is selected for semantic augmentation, the semantic labels are converted into word vectors. The similarity between these word vectors and the word vectors of the available objects to be pasted are computed. Then one of these objects is chosen from the object bank based on a criteria such as balancing the number of objects in a dataset, or adding more instances of a poor-performing object category. The chosen object is then pasted into the image in the vicinity of the most similar label.}
    \label{fig:flowchart}
\end{figure*}

An overview of SemAug is illustrated in Figure ~\ref{fig:flowchart} and detailed steps are summarized in Algorithm~\ref{algo:semaug}. In our method, we first create an object bank $\Omega$ that contains multiple instances of various objects that can be inserted into host images. The object bank can be created either from external sources such as the web, or can be created based upon an existing dataset. Due to its convenience, we opt in for the second option in this work.

\begin{wrapfigure}{L}{0.58\textwidth}
\scalebox{0.96}{
    \begin{minipage}{0.60\textwidth}
        \begin{algorithm}[H]
        \caption{\label{algo:semaug} The proposed semantic augmentation method} 
        \begin{algorithmic}[1]
        \algnewcommand\algorithmicinput{\textbf{Inputs:}}
        \algnewcommand\INPUT{\item[\algorithmicinput]}
        \algnewcommand\algorithmicoutput{\textbf{Output:}}
        \algnewcommand\OUTPUT{\item[\algorithmicoutput]} 
        \INPUT An image dataset {$\cal {D}$}
        
        \OUTPUT Semantically augmented images $\mathcal{\tilde{D}}$
        \Procedure{SemAug}{$\cal {D}$} 
        \Statex \textit{\# Object Bank Creation}
        \For {each ${\mathbf{I}}$ in $\cal {D}$}:
        \State $\mathbf{M} \gets \texttt{GetMask}(\mathbf{I})$
        
        \For {each object $k$ in $\mathbf{I}$}:
        
        \State $\mathbf{I}_c^k ,\mathbf{M}_c^k = \texttt{crop}(\mathbf{I}^k,\mathbf{M}^k |{bbox}^k)$
        \State $\mathbf{L}_e$ = \texttt{GetLanguageEmbedding}($\mathbf{L}_w(k)$)
        \State $\Omega \gets \mathbf{I}_c^k ,\mathbf{M}_c^k,\mathbf{L}_w,\mathbf{L}_e$
        \EndFor
        \EndFor
        \State\(\mathbf{D} : \mathbf{L}_w \rightarrow \mathbf{L}_e\)
        \Statex \textit{\# Image Augmentation}
        \For {each $\mathbf{I}$ in $\cal {D}$}:
            \State $a^*,b^* = \texttt{FindBestMatch}({\mathbf{I}},\Omega)$ from (\ref{eq:instance-based}) 
            
            \State $\mathbf{I}^*,\mathbf{M}^*$ = \texttt{PadZeros}($\mathbf{I},a^*,b^*$)
            
            \State $\mathcal{\tilde{D}} \gets \tilde{\mathbf{I}} = \mathbf{I} \odot (1-\mathbf{M^*}) + \mathbf{I^*}$
         
        \EndFor
        \EndProcedure
        
        \end{algorithmic}
        \end{algorithm}

    \end{minipage}
}
\end{wrapfigure}

Once the object bank is created, we explore the language representations associated with the objects in the bank and analyze them with respect to the objects appearing in each training image. By matching the high level semantics through the lens of language embeddings we identify \texttt{what} and \texttt{where} to insert from the bank to a host image. Details  are explained in this section.
 
At the end, we can apply any other kind of augmentation such as the traditional image transformations to the pipeline before passing the dataset to the training engine.

\subsubsection{Object bank creation} 
\label{sec:object_bank_creation}
To create the object bank, we first start by generating an approximate segmentation mask $\mathbf{M}$ for each image $\mathbf{I}$ in the dataset. These masks can be generated by leveraging a static side model such as a DeepLab \cite{deeplab} model (later we study the impact of mask quality and observe that even rough masks are enough). For the \(k^{th}\) object in the image, its mask can be denoted as $\mathbf{M}^k \in \{ 0,1 \} ^{W\times H}$. The mask associates a binary value where the \(k^{th}\) object appears in the image, such that \(\mathbf{M}_{x,y}^k = 1\) if the pixel at \((x,y)\) belongs to the \(k^{th}\) object. The object's masked image can be denoted as \(\mathbf{I}^k \in \mathbb{R}^{W\times H}\). Before placing the object's image and mask in the object bank, they are first cropped according to the object's bounding box to reduce their storage space:
\begin{equation}
\label{eq:objectbank}
\mathbf{I}_c^k ,\mathbf{M}_c^k = \texttt{crop}(\mathbf{I}^k,\mathbf{M}^k |{bbox}^k).
\end{equation}
This process is done once, before training, and for all images in the training dataset. The last step of the object bank creation is to create a dictionary, \(\mathbf{D}\), of all the words (or "tokens") in semantic labels, \(\mathbf{L}_w\), and their corresponding word embeddings, \(\mathbf{L}_e\), such that \(\mathbf{D} : \mathbf{L}_w \rightarrow \mathbf{L}_e\). To this end, we leverage an existing language model to extract the embedding descriptions of the semantics.

\subsubsection{Matching semantics through word embeddings}
\label{sec:matching_embeddings}
Once we obtain the language representations of objects, we perform a similarity analysis to identify a target object from the bank (\texttt{what}) and where to place it in the host image (\texttt{where}). We use the cosine similarity metric to measure the embedding similarity, however other metrics such as a Euclidean distance can be used too. To this end, let $a$  and $b$ denote two word vectors we wish to compare. 
The cosine similarity is defined as:
\begin{equation}
\label{eq:cosine}
    f_{sim}(a,b) =\dfrac{a \cdot b}{\lVert {a} \rVert \cdot  \lVert {b} \rVert} = \dfrac{\sum_{i=1}^{d} a_i b_i}{\sqrt{\sum_{i=1}^{d} a_i^2} \sqrt{\sum_{i=1}^{d} b_i^2} },
\end{equation}
where $d$ is the word embedding dimension. Supplementary materials \cite{supplementary} contain ablations on the choice of the embedding dimension.

\noindent A simple strategy for object selection is to choose the object pair with the highest similarity:
\begin{equation}
    a^*, b^* = \operatorname*{argmax}_{a\in \{\mathbf{L}_e^{\mathbf{I}}\}, b\in \{\mathbf{L}_e^{bank}\}} f_{sim}(a,b),
\end{equation}
where $\{\mathbf{L}_e^{\mathbf{I}}\}$ and $\{\mathbf{L}_e^{bank}\}$ denote all possible embedding choices within the host image and the bank, respectively, and a random instance from the $b^*$ object category will be inserted in the host image at the $a^*$ location. While this strategy intuitively might make sense, it has a down-side that during different epochs, a same object category will be selected every time. To address this issue, we choose from the top $N$ most similar embeddings, the object category with the least number of appearances so far in the current epoch. Note that the number of instance appearances is constantly being updated due to object injection and batch-wise training. Therefore, we are dynamically promoting for a better balancing of the training examples, while also choosing categories with high semantic similarity:

\begin{equation}
    \label{eq:instance-based}
    a^*, b^* = \operatorname*{argmin}_{b\in \{top~N ~sim\}}
    {count}\left(\operatorname*{arg-top~N}_{a\in \{\mathbf{L}_e^{\mathbf{I}}\}, b\in \{\mathbf{L}_e^{bank}\}} f_{sim}(a,b)\right).
\end{equation}

\begin{figure*}[t!]
    \centering
    \includegraphics[width=0.85\linewidth]{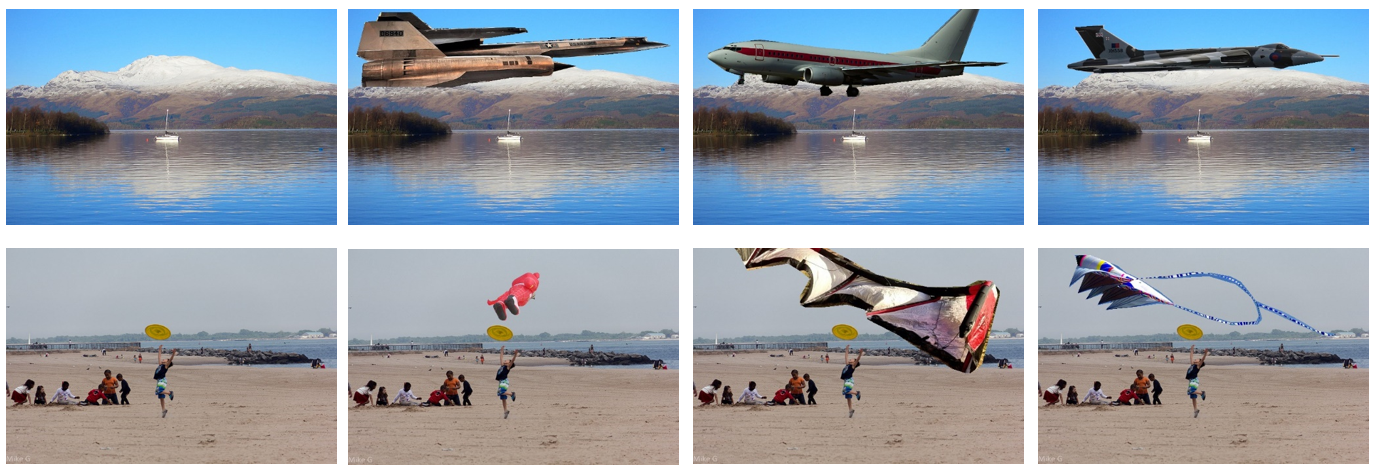}
    \caption{Our method can augment different instances from the same object category. Top row: Different instances from the category airplane are inserted. Bottom row: Different instances from the category kite are inserted. }
    \label{fig:same_category_different_instances}
\end{figure*}

\begin{figure*}[!ht]
    \centering
    \includegraphics[width=0.85\linewidth]{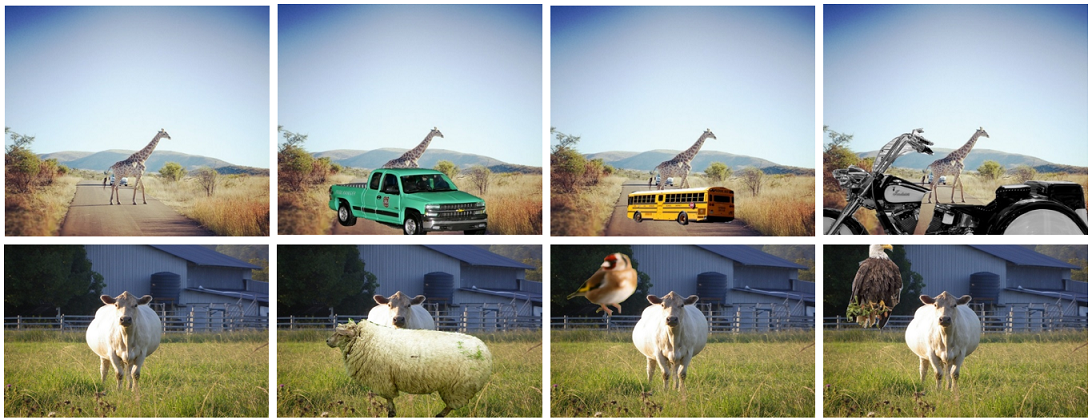}
    \caption{Our method can augment instances of different categories. Top row: An instance from the categories truck, bus and motorcycle are inserted. Bottom row: An instance from the categories sheep, and bird are inserted. Note that objects are inserted in logical locations: vehicles on roads, birds in trees, sheep on grass.}
    \label{fig:different_categories}
\end{figure*}

\subsubsection{Image augmentation}
\label{sec:image_augmentation}

Following the semantic similarity matching strategy of (\ref{eq:instance-based}), we obtain an object category $b^*$ and a target host object $a^*$, i.e., \texttt{what} and \texttt{where}. In this section, we explain the actual image augmentation procedure using the selected pairs. To this end, first, we randomly select an object-image instance of type $b^*$ from the bank (Note that different instances can be selected at different epochs, thereby presenting diversified object instances to the training algorithm). Then we scale it randomly by a factor between 5-40\% of the image $\mathbf{I}$ width. The object image is scaled using linear interpolation and the mask is scaled using nearest neighbour interpolation. 
To ensure the pasted objects are not too small or too large, we repeat the random scaling until the resized object's area falls within some bounds $(A_{min},A_{max})$. 
Next, the center coordinates of the incoming object $(x_b,y_b)$ is selected at a random vicinity of the corners of the most similar object in the image, as follows: 
\begin{align}
    \label{eq:xy_object}
    x_{b} = x_a \pm \dfrac{w_a}{2} \pm \epsilon_a, &&  y_{b} = y_a \pm \dfrac{h_a}{2} \pm \epsilon_b,
\end{align}
where $(x_a,y_a)$ is the center coordinate of the host object, $(w_a,h_a)$ are its bounding box width and height, and $\epsilon_a$ and $\epsilon_b$ are small random values to add extra randomness in the placement. If this results in occlusions, the bounding box labels are updated accordingly.

Once the center of the object to be pasted is found, its image and mask are padded with zeros to fit the shape of the training image $\mathbf{I}$. The zero-padded image and mask are denoted as $\mathbf{I^*}$ and $\mathbf{M^*}$ respectively. To compute the final augmented image and mask ($\tilde{\mathbf{I}}, \tilde{\mathbf{M}}$) the followings are used:
\begin{equation}
    \label{eq:final_image}
    \tilde{\mathbf{I}} = \mathbf{I} \odot (1-\mathbf{M^*}) + \mathbf{I^*},
\end{equation}
\begin{equation}
    \label{eq:final_mask}
    \tilde{\mathbf{M}} = \mathbf{M} \odot (1-\mathbf{M^*}) + \mathbf{M^*},
\end{equation}
where $\odot$ denotes the element-wise multiplication. At this point, the semantically augmented image $\tilde{\mathbf{I}}$ is ready to be used for training. Semantic augmentation examples are seen in Figure \ref{fig:same_category_different_instances}, where the method pasted different instances from the same category and Figure \ref{fig:different_categories} where different categories were pasted into the image. In both figures, the pasted objects are contextually relevant to the scenes. 

\subsection{Computational complexity}
In its simplest form, SemAug uses a dictionary lookup to gather the word embeddings of an image, computes a similarity metric then chooses an object to be pasted based on the similarity values. The complexity of the initial creation of the dictionary is O(len(D)) where len(D) is the number of dictionary items. To get a value from the dictionary is O(1), therefore for each image it is O(obj) where obj is the number of labeled objects in the image. For cosine similarity, the overall computational complexity is O(len(D).Obj.d), where d is the dimension of the embeddings, as the similarity is being calculated for every pair of word embeddings. This negligible overhead is the extra computation that is required to take place for each image during training. For a Mask-RCNN model with ResNet-50 backbone trained on COCO, the additional FLOPs required will be ~480,000 (80 objects$\times$20 objects$\times$300 dimension vector).
This corresponds to only 0.000107\% additional FLOPs.
Inference does not incur any extra overhead as it is unchanged.

\section{Experiments} \label{sec:experiments}
This section reviews the results of experiments in support of our method, and provides discussions around them.

\subsection{Setup}
\noindent\textbf{Architecture:} For a fair comparison with existing cut-paste methods, we used Mask R-CNN \cite{he2018mask} with ResNet \cite{he2015deep} backbone and the publicly available MMDet toolkit \cite{mmdetection} on the MS COCO dataset \cite{coco}. We also show that SemAug is compatible with Faster-RCNN \cite{ren2016faster} and RetinaNet \cite{retinanet} using this framework in addition to showing that it improves data efficiency. Otherwise, we employ an Efficientdet-d0 \cite{effdet} as the backbone for some PASCAL VOC experiments. We ran the experiments on a server equipped with eight NVIDIA V100 GPUs.  

\noindent\textbf{Training details:} For the experiments in this paper, we choose a default \textit{N} of 3, for the top 3 most similar embeddings. For $(A_{min},A_{max})$ we use the values (300, 90000). Additionally, default image resolutions from MMDet/Efficientdet config files were used  \cite{mmdetection}, \cite{effdet}.

\noindent\textbf{Datasets:} We evaluate SemAug on two standard benchmarks: MS COCO \cite{coco} and Pascal VOC \cite{voc}. The COCO dataset contains 118k training, 5k validation images, and 41k test images over 80 object categories. The Pascal dataset is considerably smaller containing only 20 object categories. Following the standard practice, we use VOC'07+12 training set (16551 examples) for training, and evaluate the models on the VOC'07 test set (4952 images). In contrast to previous object-based approaches such as \cite{cocp} and \cite{instaboost} which relied on accurate ground-truth segmentation masks, in our method these masks were generated with an off-the-shelf DeepLab-v2 \cite{deeplab} model when needed (See Section \ref{sec:results} for details).

For language grounding, we used the word embeddings from Glove \cite{glove} trained over a 2014 Wikipedia dump + Gigaword 5 \cite{glove} with a dimension of 300. 

\subsection{Results}
\label{sec:results}
\subsubsection{Comparison to cut-paste methods:} \label{sec:sota}In this subsection, we compare with state-of-the-art cut-paste methods (e.g., COCP \cite{cocp}, InstaBoost \cite{instaboost} and Context-DA \cite{context}) using Mask R-CNN based on ResNet101 on object detection and instance segmentation tasks. The results can be seen in Table \ref{tab:cocp} where our SemAug outperforms ContextDA \cite{context}, InstaBoost \cite{instaboost} and COCP \cite{cocp} by 2.8\%, 2.1\% and 1.6\% on object detection, respectively (`Vanilla' refers to traditional augmentations used by default in MMdet training pipeline, and is applied for all benchmarks). On the COCO test-dev dataset, SemAug achieves 41.6\% mAP, while Vanilla 39.4\%, and Instaboost 39.5\%. Additionally, our method sees similar performance boosts on the task of instance segmentation. Specifically, our SemAug outperforms ContextDA \cite{context}, InstaBoost \cite{instaboost}, and COCP \cite{cocp} by 2.4\%, 1.7\%, and 1.5\%, respectively. An additional comparison to Context-DA is provided in the supplementary materials. Based on these observations, our SemAug method achieves better accuracy than other cut-paste approaches. 

\begin{table*}[h]
\centering
\resizebox{\textwidth}{!}{
\begin{tabular}{lcccccccccccc}
\toprule
& \multicolumn{3}{c}{APdet, IOU} & \multicolumn{3}{c}{APdet, Area} & \multicolumn{3}{c}{APseg, IOU} & \multicolumn{3}{c}{APseg, Area}\\ \cmidrule(lr){2-4}\cmidrule(lr){5-7}\cmidrule(lr){8-10}\cmidrule(lr){11-13}
&  0.5:0.95 & 0.50 & 0.75 & Sma. & Med. & Lar. &  0.5:0.95 & 0.50 & 0.75 & Sma. & Med. & Lar.\\
\midrule
Vanilla \cite{he2018mask}  &  39.6 & 61.4 & 43.5  & 23.1 & 43.8 & 51.5 & 36.0 & 57.9 & 38.7 & 19.0 & 39.7 & 49.5\\
Context-DA  \cite{context} &  39.9 & 61.4   & 43.7  & 23.0 & 44.2 & 51.5 & 36.2 & 58.2 & 38.4 & 19.4 & 39.8 & 49.9\\
InstaBoost \cite{instaboost} &  40.6 & 62.1   & 44.3  & 24.4 & 44.6 & 53.3 & 36.8 & 58.6 & 39.6 & 20.4 & 40.4 & 50.8\\
COCP \cite{cocp} &  41.1 & 62.5   & 45.0  & 23.3 & 44.6 & 52.4 & 37.0 & 58.9 & 39.4 & 19.4 & 40.5 & 50.7\\
SemAug  &  \textbf{42.7$\pm$ 0.13}  &  \textbf{64.5}   & \textbf{46.9}  & \textbf{25.6} & \textbf{47.3} & \textbf{56.1} & \textbf{38.5$\pm$ 0.11} & \textbf{61.3} & \textbf{41.1} & \textbf{21.7} & \textbf{42.3} & \textbf{53.4}\\
\bottomrule
\end{tabular}
}
\caption{Comparison to other state-of-the-art (SOTA) methods using MMdet and Mask RCNN with a Resnet 101 backbone on COCO val. Context-DA and COCP numbers taken from the COCP paper \cite{cocp}. The APdet and APseg for SemAug are reported as the mean value and 95\% Confidence Intervals based on 5 repeat trails. }
\label{tab:cocp}
\end{table*}

\begin{table*}[]
\begin{center}
\resizebox{\textwidth}{!}{
\begin{tabular}{lcccccccccccccc}
\toprule
\multirow{2}{*}{ } & \multirow{2}{*}{Detector} & \multirow{2}{*}{Backbone} & \multicolumn{3}{c}{APdet, IOU} & \multicolumn{3}{c}{APdet, Area}& \multicolumn{3}{c}{ARdet, \#Det}& \multicolumn{3}{c}{ARdet, Area}\\ \cline{4-15} 
& & & 0.5:0.95 & 0.50 & 0.75 & Sma. & Med. & Lar. &  1 & 10 & 100 & Sma. & Med. & Lar.\\
\midrule
Vanilla & \multirow{2}{*}{Faster R-CNN \cite{ren2016faster}} & \multirow{2}{*}{Resnet-50}& 36.5 & 58.4   & 39.5  & 21.7 & 40.2 &46.8 & 30.5 & 49.3 & 51.9 & 32.8 & 56.2 & 65.2\\
SemAug& & &  \textbf{38.5}  &  \textbf{60.7}   & \textbf{41.5}  & \textbf{23.9} & \textbf{42.4} & \textbf{49.5} & \textbf{31.7} & \textbf{50.4} & \textbf{53.0} & \textbf{34.8} & \textbf{57.6} & \textbf{66.7}\\\midrule

Vanilla & \multirow{2}{*}{Faster-RCNN \cite{ren2016faster}} & \multirow{2}{*}{Resnet-101} & 38.5 & 60.3   & 41.7  & 22.7 & 42.9 & 49.7 & 31.7 & 50.6 & 53.3 & 34.8 & 57.7 & 66.9\\
SemAug& & &  \textbf{40.5}  &  \textbf{62.8}   & \textbf{44.4}  & \textbf{26.1} & \textbf{45.1} & \textbf{52.0} & \textbf{32.8} & \textbf{52.0} & \textbf{54.8} & \textbf{37.4} & \textbf{59.6} & \textbf{68.8}\\\midrule

Vanilla & \multirow{2}{*}{RetinaNet \cite{retinanet}} & \multirow{2}{*}{Resnet-50}& 35.3 & 55.2   & 37.6  & 19.4 & 39.3 &46.5 & 30.4 & 49.2 & 52.3 & 31.9 & 56.4 & 66.9\\
SemAug& & &  \textbf{37.4}  &  \textbf{57.7}   & \textbf{40.3}  & \textbf{22.3} & \textbf{41.4} & \textbf{49.5} & \textbf{31.7} & \textbf{50.4} & \textbf{53.6} & \textbf{33.5} & \textbf{58.3} & \textbf{68.6}\\\midrule

Vanilla & \multirow{2}{*}{RetinaNet \cite{retinanet}} & \multirow{2}{*}{Resnet-101}& 37.6 & 57.5   & 40.2  & 20.8 & 42.2 & 49.9 & 31.7 & 50.6 & 53.8 & 33.2 & 58.4 & 69.7\\
SemAug& & &  \textbf{39.6}  &  \textbf{60.0}   & \textbf{42.4}  & \textbf{23.4} & \textbf{44.6} & \textbf{52.3} & \textbf{32.9} & \textbf{51.9} & \textbf{55.2} & \textbf{35.5} & \textbf{60.3} & \textbf{71.1}\\\midrule

Vanilla & \multirow{2}{*}{Mask-RCNN \cite{he2018mask}} & \multirow{2}{*}{Resnet-50}& 37.8 & 59.5   & 41.0  & 23.2 & 41.4 &49.4 & 31.7 & 50.6 & 53.3 & 35.1 & 57.5 & 66.8\\
SemAug& & &  \textbf{39.2}  &  \textbf{61.4}   & \textbf{42.9}  & \textbf{24.8} & \textbf{43.2} & \textbf{50.9} & \textbf{32.2} & \textbf{51.2} & \textbf{53.9} & \textbf{35.8} & \textbf{58.2} & \textbf{68.1}\\\midrule

Vanilla & \multirow{2}{*}{Mask-RCNN \cite{he2018mask}} & \multirow{2}{*}{Resnet-101}& 39.6 & 61.4   & 43.5  & 23.1 & 43.8 & 51.5 & 32.3 & 51.5 & 54.2 & 34.9 & 58.8 & 68.5\\
SemAug& & &  \textbf{42.7}  &  \textbf{64.5}   & \textbf{46.9}  & \textbf{25.6} & \textbf{47.3} & \textbf{56.1} & \textbf{34.2} & \textbf{54.4} & \textbf{57.3} & \textbf{38.7} & \textbf{62.1} & \textbf{71.8}\\
\bottomrule
\end{tabular}
}
\end{center}
\caption{Object detection results (\%) on the COCO val benchmark with different size backbones and default parameters.} 
\label{tab:backbone}
\end{table*}

\subsubsection{Results of incorporating SemAug in labeled datasets and different architectures:}  Our SemAug method has been shown to work on a variety of state-of-the-art object detection architectures with different capacities as shown in Table \ref{tab:backbone}. This exemplifies how our augmentation strategy considers context without the training and inference overhead of an additional context models allowing for easy adoption into existing models. 

\subsubsection{Results using smaller dataset sizes:} 
\begin{SCfigure}
\centering\includegraphics[width=0.40\textwidth]{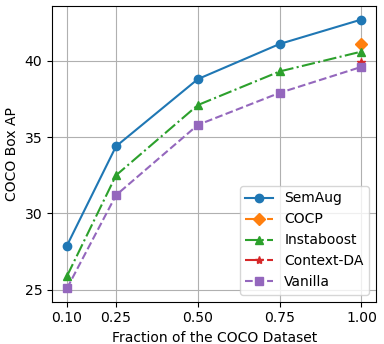}
\caption{Data-efficiency on the COCO val benchmark using Mask-RCNN with a Resnet-101 backbone. The results show a consistent increase of $\approx3\%$ mAP over vanilla in both the low data and high data regimes. Curves (fractional results) are shown for methods for which code was available and could run.}

\label{fig:efficiency}
\end{SCfigure}

In many real-world applications, it is difficult to collect and label data. Therefore, we evaluated the performance of our method in settings where less labeled data was available. As shown in Figure \ref{fig:efficiency}, SemAug was able to provide a boost in performance even in the low data regimes using a fraction of the COCO dataset.

\begin{figure}
    \centering
    \includegraphics[width=0.70\linewidth]{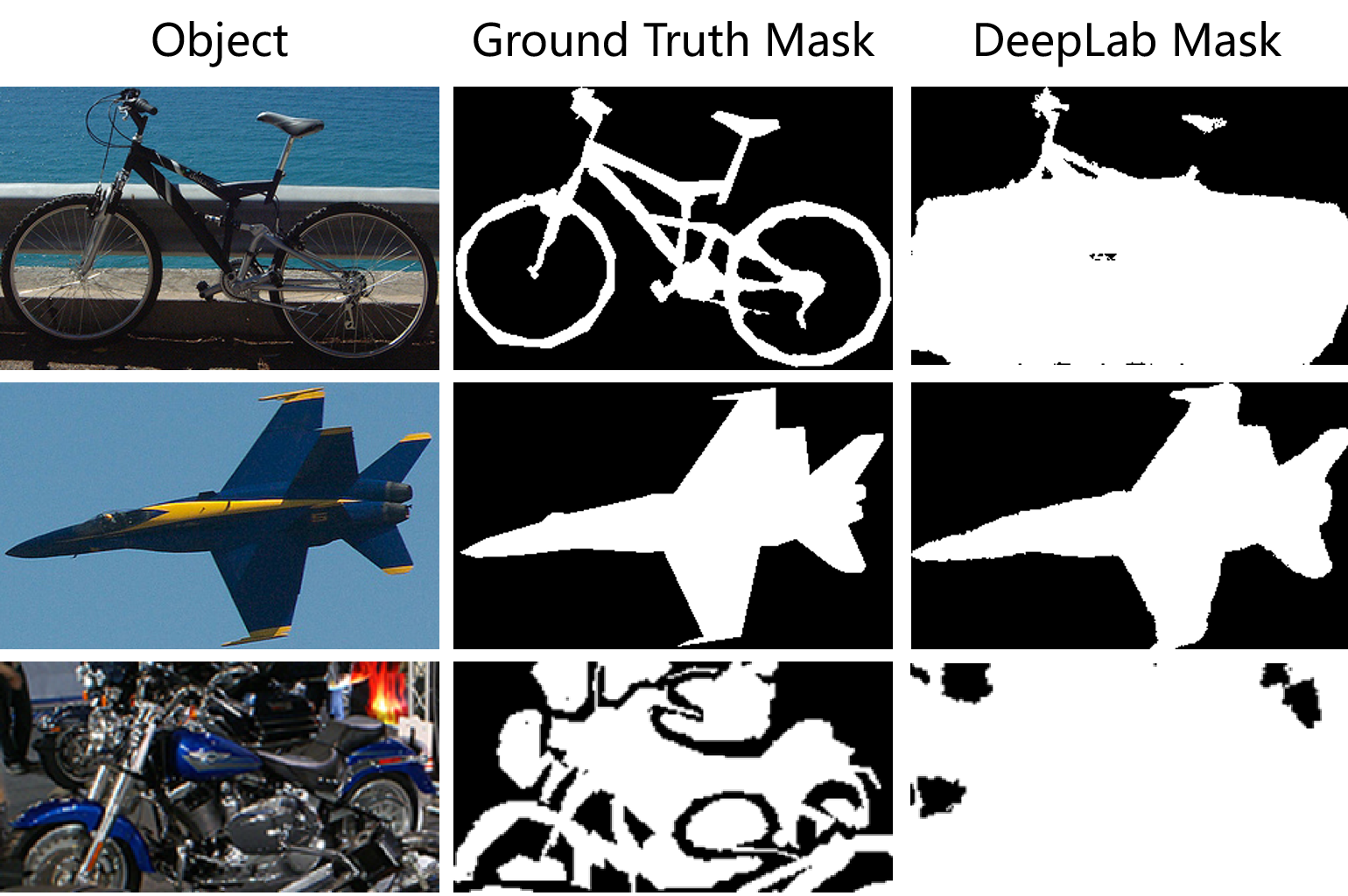}
    \caption{Mask quality examples. Ground truth masks are much more precise than DeepLab masks.}
    \label{fig:masks}
\end{figure}

\subsubsection{Ablation of object bank size and mask qualities:} 

As mentioned in Section \ref{sec:object_bank_creation}, to create an object bank from images without given masks, we can use an off-the-shelf model for convenience such as deeplab. Note the deeplab generated masks are only being used for object bank creation, and therefore the algorithm is not sensitive to their quality. In the case of a bad quality mask (larger or smaller mask) the impact will be similar to either adding an object with more context, or an occluded (partial) object, which may in fact improve the generalization. As shown in Figure \ref{fig:masks}, the deeplab masks were less precise than the ground truth masks which were only provided for a small subset (VOC-seg, 1464 images) of the VOC dataset. For this experiment we used Efficientdet-d0 with deeplab masks on VOC-seg as well as the whole Pascal VOC training dataset. As shown in Table \ref{tab:semisuper}, our method is not sensitive to the quality of the masks in the object bank. An added benefit to using deeplab masks is the ability to supplement the object bank with additional objects from previously unlabeled images. In this regard, we observe that while the deeplab masks were worse quality than the ground truth masks, they provided better performance when additional objects were added to the object bank.

\begin{table}
\centering
\begin{tabular}{p{0.15\linewidth} p{0.25\linewidth} p{0.25\linewidth} p{0.15\linewidth}}
\toprule
Method & Bank Dataset & Object Mask & mAP\\
\midrule
Vanilla & --- & --- & 73.59 \\ 
SemAug & VOC-seg & DeepLab & 77.19 \\
SemAug & VOC-seg & Ground Truth & 77.31 \\
SemAug & VOC-all & DeepLab & \textbf{77.35} \\
\bottomrule
\end{tabular}
\caption{Effect of object bank mask quality on SemAug.}
\label{tab:semisuper}
\end{table}

\subsubsection{Results of adding new categories:} The ability to add new categories to datasets is applicable to important real-world scenarios when target categories are rare or uncommon in nature (e.g. detection of security threats). Due to the inherent constraints of previous works \cite{instaboost}, \cite{cocp}, \cite{context}, they are unable to add new categories in a knowledgeable manner.  To demonstrate SemAug's ability to add new categories, we use Efficientdet-d0 on the Pascal VOC dataset and remove all images of a specific category. Object instances from that category are then pasted into the remaining images during training where appropriate. For this experiment, we chose an \textit{N} of 5, and only paste objects from the removed category if they are in the top 5 most similar embeddings. We compare the average AP of the other 19 categories before and after the addition of a new category to show that it does not harm the other categories. We do this experiment on the first five categories of VOC, one at a time. Results are shown in Table \ref{tab:addcat}. As observed, SemAug is able to add categories with decent results, while not harming the detection of existing categories in the dataset.

\begin{table}
\begin{center}
\setlength{\tabcolsep}{4pt}
\begin{tabular}{cccccc}
\toprule
 & Aeroplane & Bicycle & Bird & Boat & Bottle \\
\midrule
Categories before & 79.3 & 79.5 & 79.4 & 80.7 & 81.1  \\
Categories after & 79.6 & 79.7 & 79.6 & 80.6 & 80.9 \\
\midrule
New category & 64.6 & 78.2 & 67.4 & 39.9 & 37.9  \\
\bottomrule
\end{tabular}
\end{center}

\caption{Effect of removing categories from the PASCAL VOC dataset then adding them back using semantic augmentation.Top two rows are the mAP results of all categories except the newly augmented category. Results are recorded as mAP.} 
\label{tab:addcat}
\end{table}

\subsubsection{Additional comparisons on Pascal VOC:}
In section \ref{sec:sota}, we compared SemAug with several other SOTA methods on the COCO dataset. Here, we additionally provide a comparison with other augmentation methods on the Pascal VOC object detection task. 
As in previous papers \cite{cutmix,cocp}, we employ a Faster-RCNN network with a Resnet-50 backbone. The results are given in Table \ref{tab:cutmix}. 
In this table, Random Paste (pasting random objects at random locations) and Co-occurrence (where we paste objects based on how often they appears together in a same image) are two naive object-based augmentation approaches that are included as additional ablation results to our method.
As mentioned previously, context is important for the object selection strategy in cut and paste methods. As we can see, methods which do not consider context either degrade performance or marginally improve it; whereas, the three methods that consider context improve performance the most.

\begin{SCtable}
\setlength{\tabcolsep}{8pt}
\begin{tabular}{cc}
\toprule
Augmentation Method  & mAP\\
\midrule
Baseline &  75.6 \\
Mixup* \cite{mixup} & 73.9 (\textcolor{red}{-1.7})\\
Cutout* \cite{cutout} &  75.0 (\textcolor{red}{-0.6})\\
Random Paste  &  75.9 (\textcolor{green}{+0.3})\\
CutMix* \cite{cutmix} &  76.7 (\textcolor{green}{+1.1})\\
COCP* \cite{cocp} & 77.4 (\textcolor{green}{+1.8})\\
Co-occurrence & 79.3 (\textcolor{green}{+3.7})\\
SemAug  & \textbf{80.7} (\textcolor{green}{+5.1})\\
\bottomrule
\end{tabular}
\caption{\protect\rule{0ex}{10ex} Comparison to other augmentation methods on the Pascal VOC dataset using Faster-RCNN and a Resnet-50 backbone. * Results taken from \cite{cutmix} and \cite{cocp}.}
\label{tab:cutmix}
\end{SCtable}

\subsubsection{Ablation of the effect of scaling objects:} In this experiment, we compare the use of different scaling ranges with our method. This experiment was conducted using MMDET and the Pascal VOC dataset. Inserting an object can occlude other objects in the scene, and adding an object that is large may remove context from the image. As can be seen in Table \ref{tab:scale}, it is advantageous to scale the objects so that they are not too small as to be unrecognizable, but also not too big to be occluding other objects. 

\begin{SCtable}
    \setlength{\tabcolsep}{12pt}
    \begin{tabular}{cc}
    \toprule
    Scaling Range (\%)  & mAP \\
    \midrule
    No scaling & 79.9 \\
    5-40 & \textbf{80.7} \\
    10-30 &  80.0\\
    15-40 & 80.4 \\
    \bottomrule
    \end{tabular}
    \caption{Effect of the scaling range for objects pasted into the scene on the Pascal VOC dataset using Faster-RCNN with a Resnet 50 backbone. The objects are randomly scaled to a percentage of the image into which they are being pasted.}
    
    \label{tab:scale}
\end{SCtable}

 \subsubsection{Ablation on the object similarity metric:} In this experiment, we study the impact of object similarity methods discussed in the paper. We employ an Efficientdet-d0 \cite{effdet} as the backbone, train using the VOC'07+12 training set, and evaluate the models on the VOC'07 test set. As can be seen in the results of Table \ref{tab:objectsimilarity}, both euclidean distance and cosine similarity provide comparable results. As cosine similarity provided marginally better results, it was used as default in the paper.

\begin{SCtable}
\setlength{\tabcolsep}{12pt}
\begin{tabular}{cc}
\toprule
Object Similarity Method  & AP50\\
\midrule
Euclidean Distance &  77.16 \\
Cosine Similarity & \textbf{77.35}\\
\bottomrule
\end{tabular}
\caption{\protect\rule{0ex}{5ex} Effect of object similarity calculation choice on Pascal VOC. }
\label{tab:objectsimilarity}
\end{SCtable}

\subsection{Limitations}
\noindent As with any method, there are several limitations to the method presented. Firstly, this method uses pre-exisiting open-source word embeddings. Though this is not a core part of our method and one could choose to train their own word embeddings if necessary. Additionally, the quality of the word embeddings is related to the corpus used for training, therefore care should be taken to ensure meaningful semantic correlations exist before using for augmentation. For example, using a news-based corpora could align 'apple' more with technology than fruit. As several high quality pre-existing open-source word embeddings currently exist, this should not pose a major issue to anyone wishing to use this method. A future works section is discussed in Supplementary \cite{supplementary}.

\section{Conclusion} \label{sec:conclusion}
This paper proposes an effective technique for image augmentation by injecting contextually meaningful knowledge into training examples. Our object-level augmentation method identifies the most suitable object instances to be pasted into host images, and chooses appropriate target regions. We do that, by analyzing and matching objects and target regions through the lens of high level natural language. Our method results in consistent generalization improvements on various object detection benchmarks.

\clearpage
%
%
\bibliographystyle{splncs04}
\bibliography{references.bib}

\clearpage
\section{Supplementary Materials} \label{sec:supplementary}

This document provides supplementary materials including additional comparisons to state-of-the-art methods, experimental ablation study results, implementation details and example augmented images. The materials are delivered in the following sections:

\begin{itemize}
\item 6.1 Code
\item 6.2 Additional comparison to Context-DA
\item 6.3 Additional ablation results
\item 6.4 Additional implementation details
\item 6.5 Future Work
\item 6.6 Additional visualizations
\end{itemize}

\subsection{Code}
To facilitate the reproducibility and easier usage of our method, we release the code implementation of SemAug at

    \href{hyphens}{https://developer.huaweicloud.com/develop/aigallery/notebook/detail?id=4d9fc5b8-7fda-4b95-91c2-27deaa2c8490}. 

Additionally, a ReadMe file on documentation of how to use the code is also included in this link.

\subsection{Additional comparison to Context-DA}
In the experiments section of the paper, we compared our SemAug method with several other state-of-the-art (SOTA) methods on the COCO dataset. Here, we additionally provide a comparison with a SOTA context-based method Context-DA \cite{context} on the Pascal VOC object detection. We use the same experiment settings as \cite{context} so we can directly compare with their provided results. Baseline is a blitznet \cite{blitznet} model as the base network with vanilla augmentations. The results are given in Table \ref{tab:context}. As shown, we see a +1.9\% mAP improvement over Context-Aug \cite{context} and a +3.8\% mAP improvement over baseline \cite{blitznet}.

\begin{table*}[!b]
\centering\settowidth\rotheadsize{\bfseries(aeroplane)}
\resizebox{\textwidth}{!}
{
    \begin{tabular}{l |*{20}{c}|{c}}
        \toprule
        {Method} & 
        \rotcell{Aeroplane} &
        \rotcell{Bicycle} &
        \rotcell{Bird} &
        \rotcell{Boat} &
        \rotcell{Bottle} &
        \rotcell{Bus} &
        \rotcell{Car} &
        \rotcell{Cat} &
        \rotcell{Chair} &
        \rotcell{Cow} &
        \rotcell{Dining table} &
        \rotcell{Dog} &
        \rotcell{Horse} &
        \rotcell{Motorbike} &
        \rotcell{Person} &
        \rotcell{Potted plant} &
        \rotcell{Sheep} &
        \rotcell{Sofa} &
        \rotcell{Train} &
        \rotcell{TV Monitor} &
        \rotcell{Average}\\
        \hline
        \eqmakebox[AB][l]{Baseline} & 63.6 & 73.3 & 63.2 & 57.0 & 31.5 & 76.0 & 71.5 & 79.9 & 40.0 & 71.6 & 61.4 & 74.6 & 80.9 & 70.4 & 67.9 & 36.5 & 64.9 & 63.0 & \textbf{79.3} & 64.7 & 64.6\\
        \eqmakebox[AB][l]{Context-DA} & 69.9 & 73.8 & 63.9 & \textbf{62.6} & 35.3 & \textbf{78.3} & 73.5 & 80.6 & 42.8 & 73.8 & 62.7 & 74.5 & \textbf{81.1} & \textbf{73.2} & 68.9 & 38.1 & 67.8 & 64.3 & \textbf{79.3} & 66.1 & 66.5\\
        \eqmakebox[AB][l]{SemAug} & \textbf{70.8} & \textbf{75.6} & \textbf{68.2} & 59.3 & \textbf{41.2} & 78.1 & \textbf{78.7} & \textbf{81.5} & \textbf{45.7} & \textbf{76.2} & \textbf{68.0} & \textbf{75.3} & \textbf{81.1} & 71.8 & \textbf{71.1} & \textbf{45.0} & \textbf{69.3} & \textbf{65.4} & 79.2 & \textbf{66.6} & \textbf{68.4}\\
        \bottomrule
    \end{tabular}
}

\caption{Comparison of detection accuracy on VOC07-test. The model is trained on all categories at the same time, by using the 1464 images from VOC12train-seg and Blitznet. The first column specifies the augmentation method used in the experiment. The numbers represent average precision per class in \%. }

\label{tab:context}
\end{table*}

\subsection{Additional ablation results}

\paragraph{Ablation of object selection strategy\\} 

We include an additional baseline strategy in our ablation studies. This method is an accuracy based mechanism, where we aim to boost the performance for the object category with lowest per-object AP (Average Precision) from the top $N$ most similar object categories. This promotes to push up the lowest AP, and in turn the mAP (mean Average Precision). 

Table \ref{tab:objectselection} shows the results of this experiment for variants of our context based method. The \texttt{MostSimilar} baseline selects the most similar object category based on the cosine similarity. The \texttt{Instance} method (our default) first narrows down the selection to the top 3 most similar object categories by cosine similarity, then selects the object category with the least amount of instances in the dataset in order to mitigate the effect of unbalanced datasets, while still allowing for semantic knowledge to be injected. The \texttt{Baseline-mAP} method first narrows down the selection to the top 3 most similar object categories by cosine similarity, then selects the object category with the lowest mAP when trained using vanilla augmentation in an attempt to boost low performing categories, while still allowing for semantic knowledge to be injected. Results in Table \ref{tab:objectselection} show that these methods are comparable.

\begin{SCtable}
\setlength{\tabcolsep}{12pt}
\begin{tabular}{cc}
\toprule
Object Selection Method  & AP50\\
\midrule
\texttt{MostSimilar} &  76.66 \\
\texttt{Baseline-mAP} & 76.86\\
\texttt{Instance} &  \textbf{77.35} \\
\bottomrule
\end{tabular}
\caption{\protect\rule{0ex}{5ex}Effect of object selection strategy. Using Efficientdet-d0 on the Pascal VOC  dataset.}
\label{tab:objectselection}
\end{SCtable}

\paragraph{Ablation on averaging similarities\\} 
In this experiment, we study the impact of averaging the similarities across all objects present in the image. This was meant to encompass more of the scene as a whole as opposed to matching objects individually. As shown in Table \ref{tab:avg}, this method did not improve the results but rather degraded the performance. This may be due to the fact that not all objects in a scene are semantically related and therefore averaging the similarities does not aid in finding contextually meaningful objects to be pasted. 

\begin{SCtable}
\setlength{\tabcolsep}{12pt}
    \begin{tabular}{lcc}
    \toprule
    Method  & APdet & APseg\\
    \midrule
    Average similarities & 41.7 & 37.7\\
    No averaging & \textbf{42.7} & \textbf{38.5} \\
    \bottomrule
    \end{tabular}
    \caption{Effect of averaging similarities across all objects in the image. Performed on the COCO dataset using Mask-RCNN with a Resnet-101 backbone.}
    
    \label{tab:avg}
\end{SCtable}

\paragraph{Ablation of word embedding size\\} We compare our method using various dimensions for the GloVe pre-trained word embeddings. This allows us to see how our results change depending on the size of the embedding. Large dimensions are needed to fully capture the essence of words for more complex NLP problems such as captioning images or answering questions, but Table \ref{tab:embeddingsize} suggests that we can take advantage of the faster computation times of smaller dimensions as embedding the similarity of objects does not necessarily need larger dimensions.

\begin{table*}
\resizebox{\textwidth}{!}{
\begin{tabular}{c|ccc|ccc|ccc|ccc}
\toprule
\multirow{2}{*}{{\shortstack[c]{Word Embedding\\ Dimension}}} & \multicolumn{3}{c|}{APdet, IOU} & \multicolumn{3}{c|}{APdet, Area}& \multicolumn{3}{c|}{APseg, IOU}& \multicolumn{3}{c}{APseg, Area}\\\cline{2-13}
& 0.5:0.95 & 0.50 & 0.75 & Sma. & Med. & Lar. &  0.5:0.95 & 0.50 & 0.75 & Sma. & Med. & Lar.\\
\hline
100 & \textbf{42.7} & \textbf{64.5}   & 46.8  & \textbf{26.2} & 47.0 & \textbf{56.1} & \textbf{38.5} & 61.1 & \textbf{41.5} & \textbf{21.8} & 42.2 & 53.0\\
200 & 42.3 & 64.1   & 46.2  & 25.8 & 46.4 & 56.0 & 38.2 & 60.7 & 40.9 & 21.6 & 41.9 & 52.9\\
300  &  \textbf{42.7}  &  \textbf{64.5}   & \textbf{46.9}  & 25.6 & \textbf{47.3} & \textbf{56.1} & \textbf{38.5} & \textbf{61.3} & 41.1 & 21.7 & \textbf{42.3} & \textbf{53.4}\\
\bottomrule
\end{tabular}
}
\caption{Effect of word embedding dimension. All experiments were done using our SemAug with MMdet and Mask RCNN with a Resnet 101 backbone on COCO. Glove pretrained embeddings on the Wikipedia 2014 + Gigaword 5 dataset were used with different dimensions.}
\label{tab:embeddingsize}

\end{table*}

\paragraph{Ablation of the number of categories used for top-N\\} We compare our method using various numbers of categories for the top-N calculation. Allowing the method to choose between more options than just the single most similar object allows for similar objects with smaller representation in the dataset to be chosen. This aids in generalization as seen by the increase in AP between N=1 and N=3 in Table \ref{tab:topN}. As we used COCO, which has 80 categories, the AP appears to plateau as the N is increased. However, this number should be chosen carefully, as a smaller dataset such as VOC with only 20 categories might be forced to incorporate dissimilar categories if the N was chosen to be half the dataset. 

\begin{SCtable}
\setlength{\tabcolsep}{12pt}
    \begin{tabular}{ccc}
    \toprule
    N  & APdet & APseg\\
    \midrule
    1 & 41.6 & 37.6\\
    2 & 42.3 & 38.1\\
    3 & \textbf{42.7} & \textbf{38.5}\\
    4 & 42.4 &  38.3 \\
    5 & 42.6 & 38.4\\
    10 & \textbf{42.7} & \textbf{38.5} \\
    \bottomrule
    \end{tabular}
    \caption{\protect\rule{0ex}{8ex}Effect of N for top-N on COCO dataset using Mask-RCNN with a Resnet-101 backbone on the dataset.}
    
    \label{tab:topN}
\end{SCtable}

\paragraph{Ablation of the number of objects pasted in the image\\} We compare the use of our method to paste one or two objects into an image. This experiment was conducted using MMDET and the Pascal VOC dataset. Inserting an object can occlude other objects in the scene, and adding too many may remove context from the image. As observed in Table \ref{tab:multobj}, inserting more objects starts to hurt the performance. 

\begin{SCtable}
\setlength{\tabcolsep}{12pt}
    \begin{tabular}{cc}
    \toprule
    Number of objects  & mAP \\
    \midrule
    1 & \textbf{80.7} \\
    2 &  79.5\\
    \bottomrule
    \end{tabular}
    \caption{Effect of single or multiple objects pasted into the scene on the Pascal VOC dataset using Faster-RCNN with a Resnet 50 backbone.}
    
    \label{tab:multobj}
\end{SCtable}

\paragraph{Ablation of the effect of blending techniques\\} In this experiment, we compare the use of different blending methods with SemAug. This experiment was conducted using MMDET and the Pascal VOC dataset. Both Gaussian and averaging filters used a [5,5] kernel. Blending objects in the scene can make them appear more realistic from a human perspective, but from the results in Table \ref{tab:blur}, it does not appear to improve the network's performance.  

\begin{SCtable}
\setlength{\tabcolsep}{12pt}
    \begin{tabular}{cc}
    \toprule
    Blending Method  & mAP \\
    \midrule
    No blending & \textbf{80.7} \\
    Gaussian & 79.4 \\
    Averaging &  79.9\\
    \bottomrule
    \end{tabular}
    \caption{Effect of blending techniques for objects pasted into the scene on the Pascal VOC dataset using Faster-RCNN with a Resnet 50 backbone.}
    
    \label{tab:blur}
\end{SCtable}

\subsection{Additional implementation details}

As mentioned in the paper, not all semantic labels provided with COCO and VOC were found in the GloVe dataset. As such, the most similar word in the GloVe dataset was found manually and the word embedding for that word was used instead. Table \ref{tab:substitutions} lists these substitutions.

\begin{SCtable}
\setlength{\tabcolsep}{12pt}
{
\begin{tabular}{cc}
\toprule
Original Word & Substituted Word \\
\midrule
baseball bat & baseball \\
baseball glove & baseball \\
dining table & table  \\
fire hydrant & hydrant  \\
parking meter & parking \\
playing field & field  \\
potted plant & plant  \\
tennis racket & racket  \\
traffic light & stoplight  \\
stop sign & stoplight \\
waterdrops & droplets  \\
\bottomrule
\end{tabular}
}
\caption{\protect\rule{0ex}{20ex}Substitutions used for semantic labels which were not in the GloVe dataset.}
\label{tab:substitutions}
\end{SCtable}

\subsection{Future Works}
\noindent This work lends itself to ideas not in the scope of this paper. For example, an interesting direction would be weakly-supervised detection, where supervision comes only from a pre-built bank
of objects. Additionally, while COCO and Pascal VOC were used here, evaluation on highly imbalanced datasets and larger datasets such as OpenImages would also be merited. Lastly, it would be interesting to see how this image augmentation technique would fair on a non-CNN such as a visual transformer, or in the case of visual question answering such as in \cite{gokhale-etal-2020-mutant}.

\subsection{Additional visualizations}
\begin{figure*}
    \centering
    \includegraphics[width=0.9\linewidth]{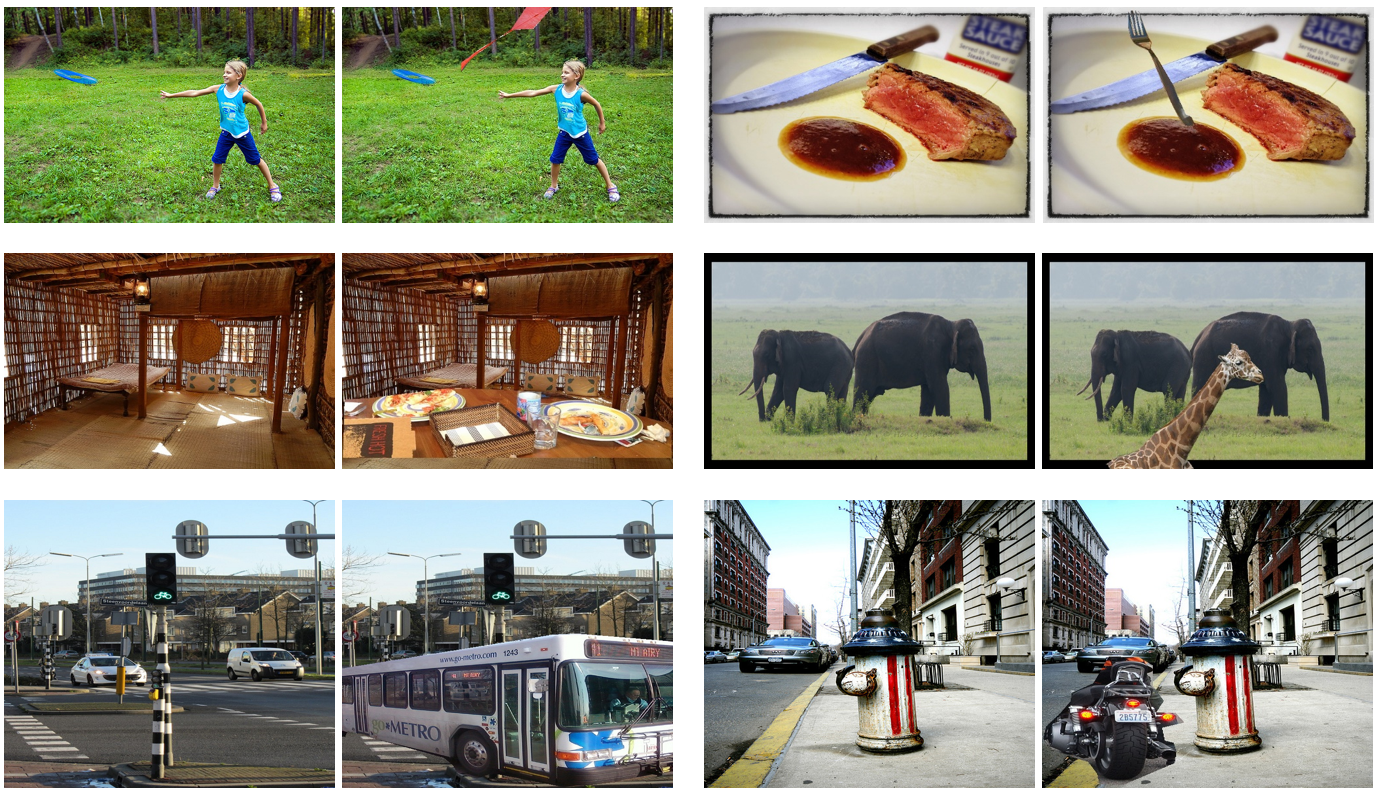}
    \caption{Examples of original (left) vs semantically augmented (right) images.}
    \label{fig:original_vs_augmented_1}
\end{figure*}

\begin{figure*}
    \centering
    \includegraphics[width=0.9\linewidth]{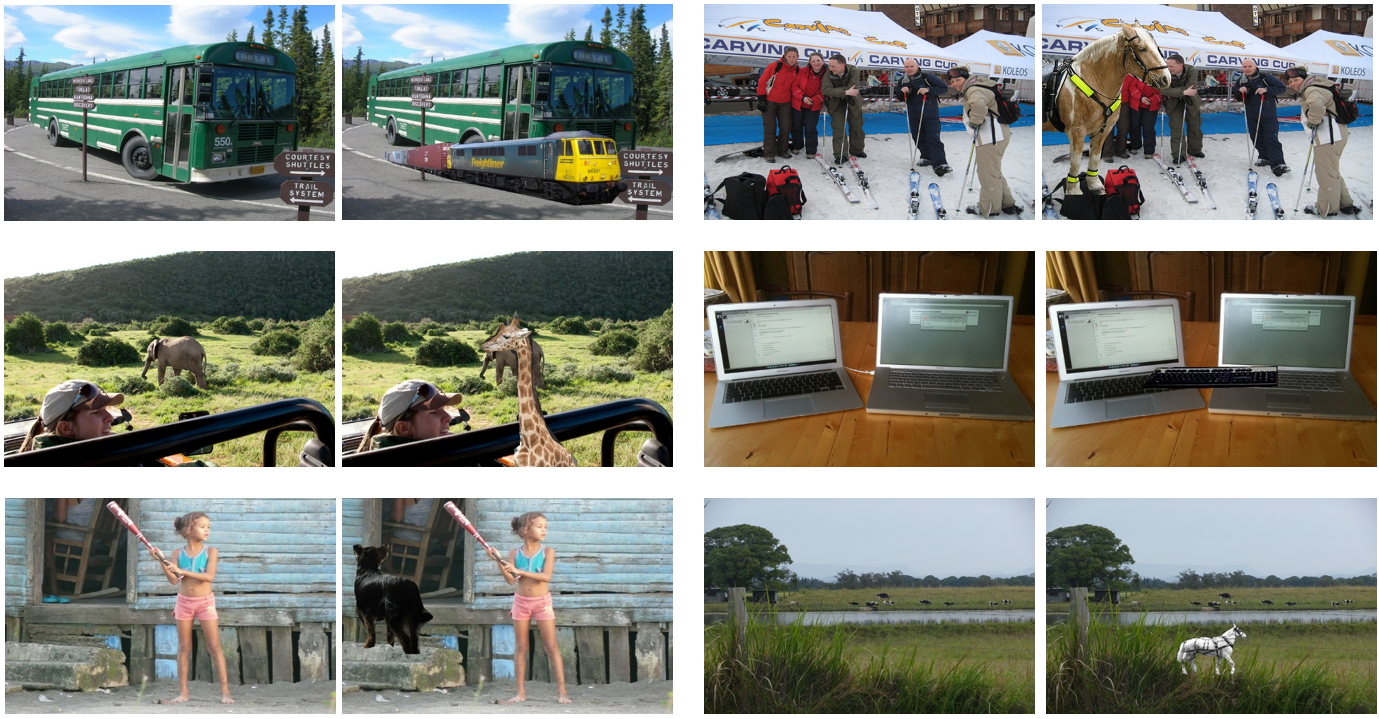}
    \caption{Examples of original (left) vs semantically augmented (right) images.}
    \label{fig:original_vs_augmented_2}
\end{figure*}

\begin{figure*}
    \centering
    \includegraphics[width=0.9\linewidth]{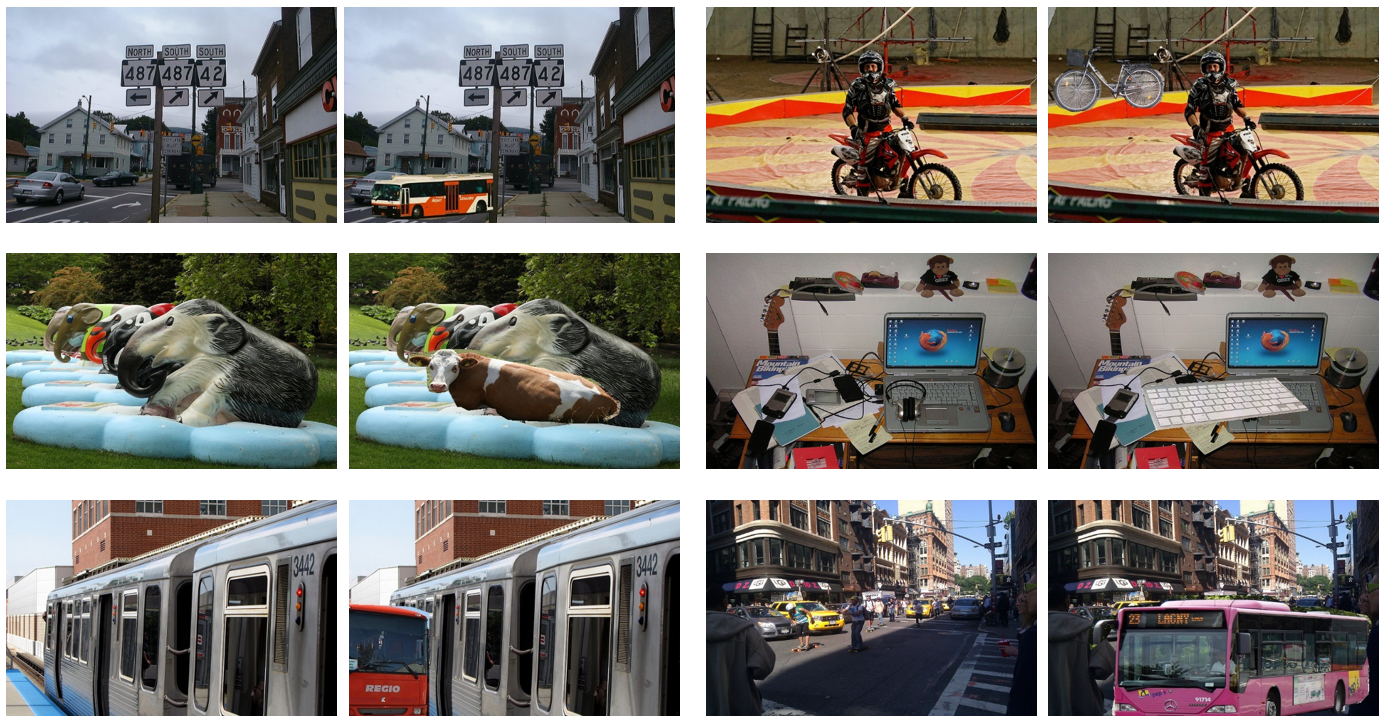}
    \caption{Examples of original (left) vs semantically augmented (right) images.}
    \label{fig:original_vs_augmented_3}
\end{figure*}

\begin{figure*}
    \centering
    \includegraphics[width=0.9\linewidth]{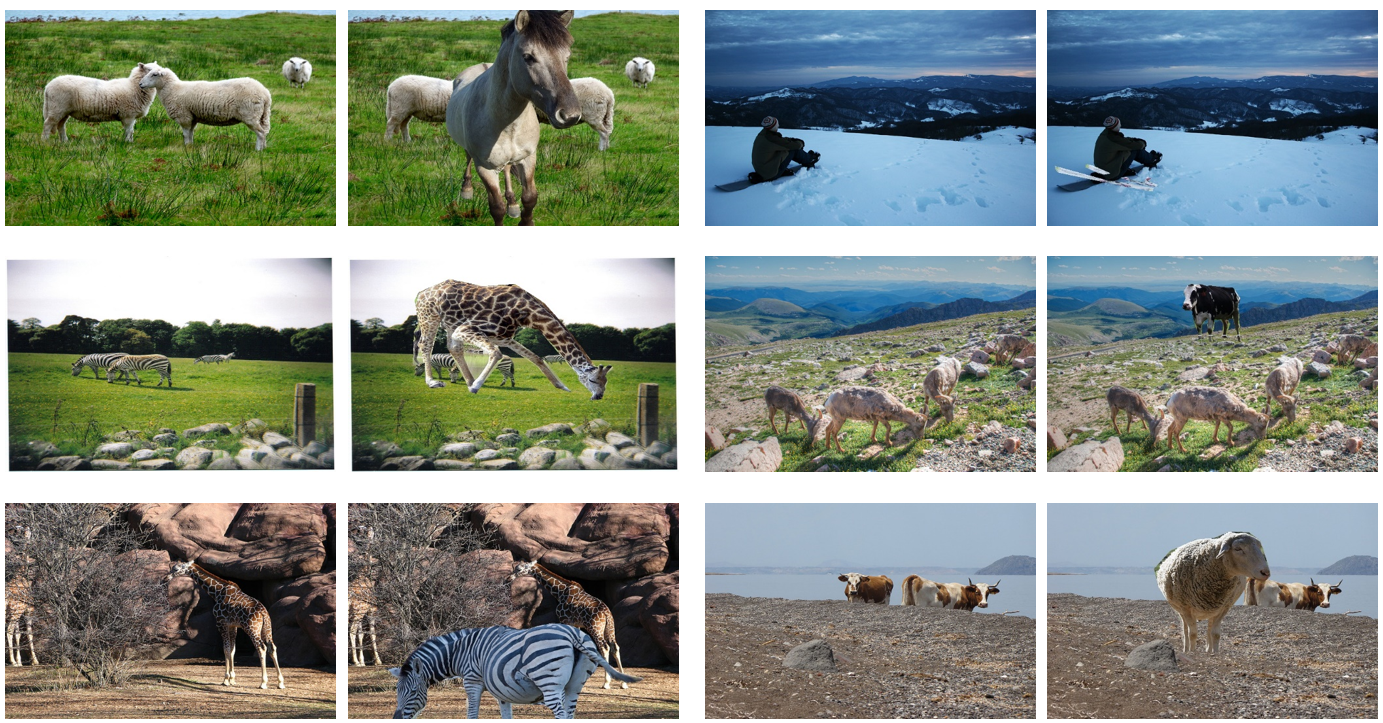}
    \caption{Examples of original (left) vs semantically augmented (right) images.}
    \label{fig:original_vs_augmented_4}
\end{figure*}

\begin{figure*}
    \centering
    \includegraphics[width=0.9\linewidth]{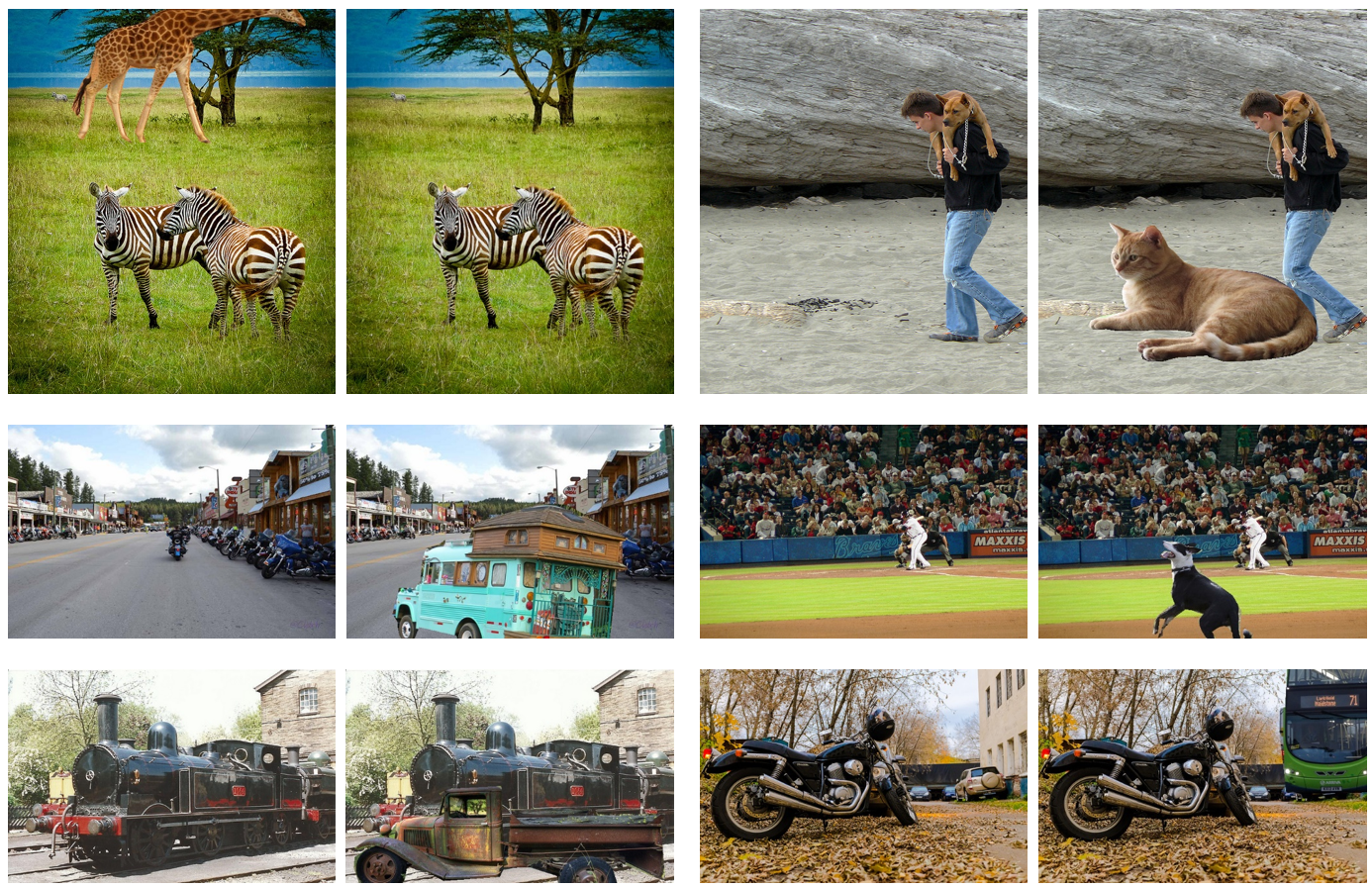}
    \caption{Examples of original (left) vs semantically augmented (right) images.}
    \label{fig:original_vs_augmented_5}
\end{figure*}

\begin{figure*}
    \centering
    \includegraphics[width=0.9\linewidth]{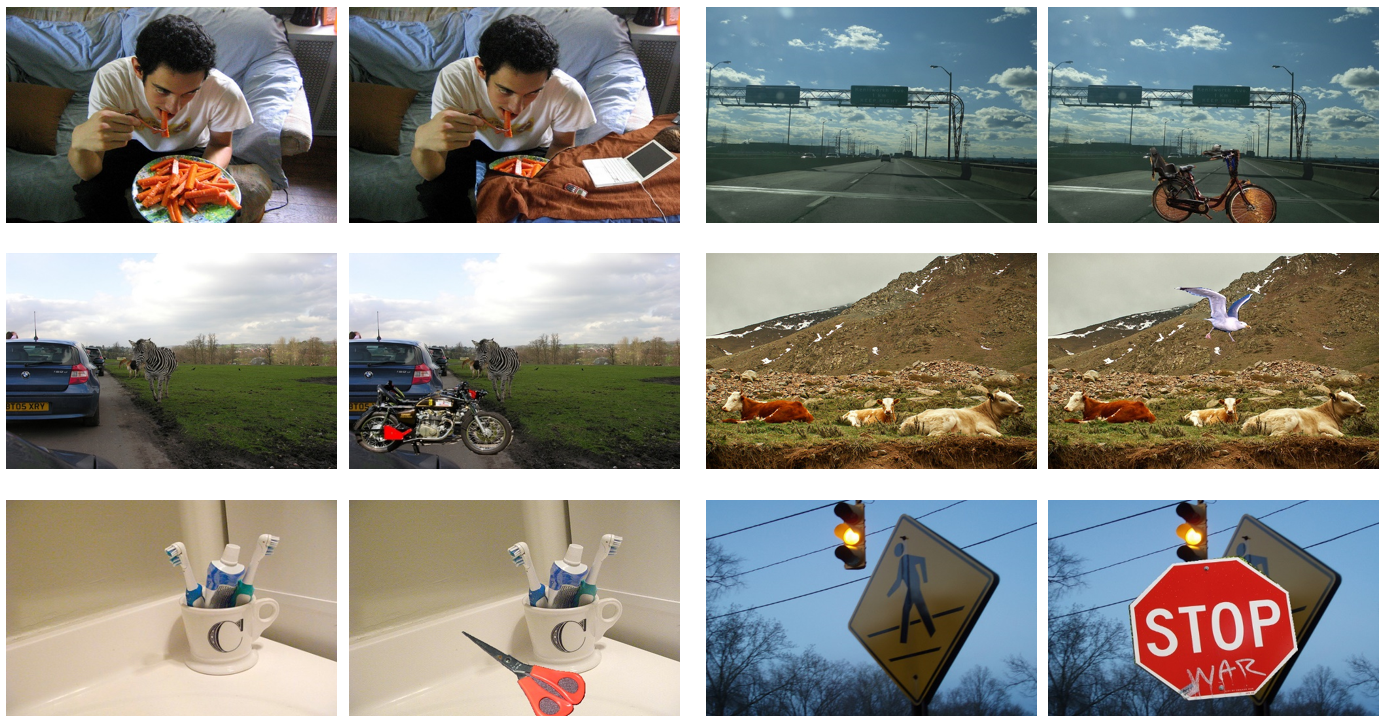}
    \caption{Examples of original (left) vs semantically augmented (right) images.}
    \label{fig:original_vs_augmented_6}
\end{figure*}

\begin{figure*}
    \centering
    \includegraphics[width=0.9\linewidth]{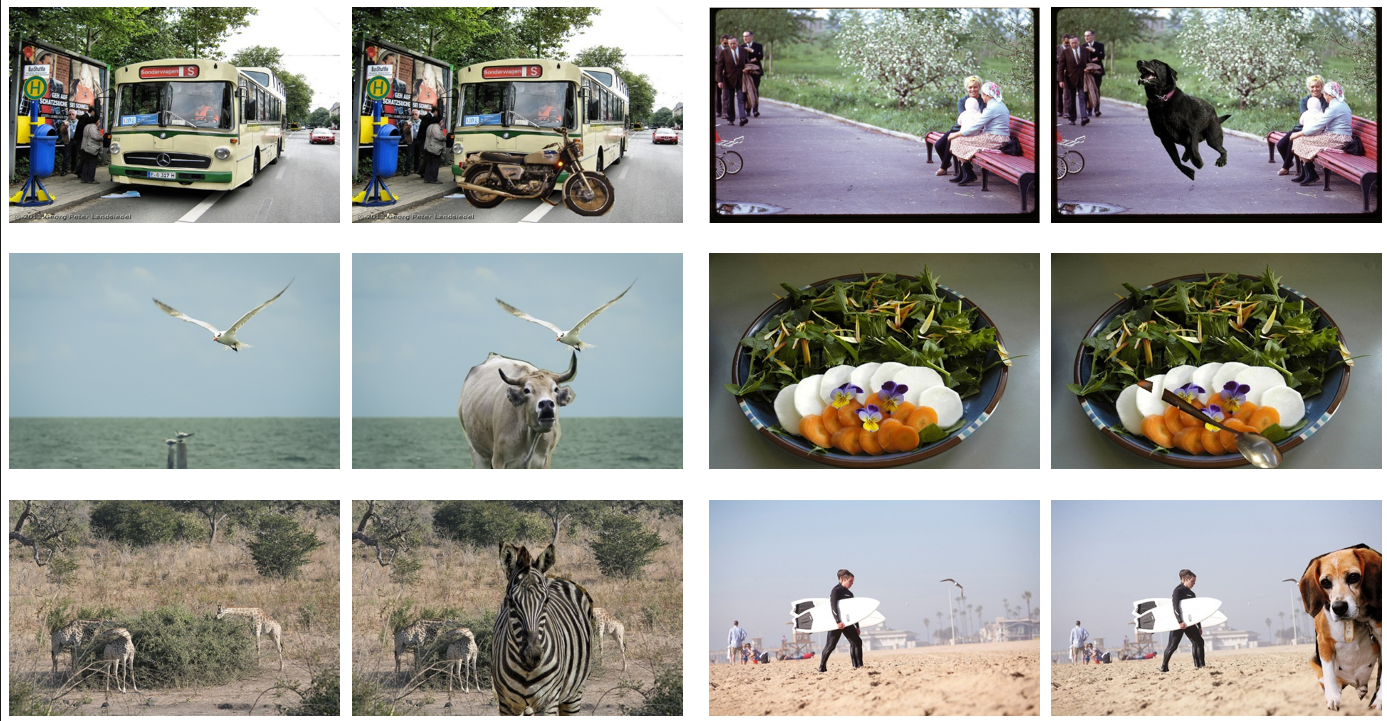}
    \caption{Examples of original (left) vs semantically augmented (right) images.}
    \label{fig:original_vs_augmented_7}
\end{figure*}

\begin{figure*}
    \centering
    \includegraphics[width=0.9\linewidth]{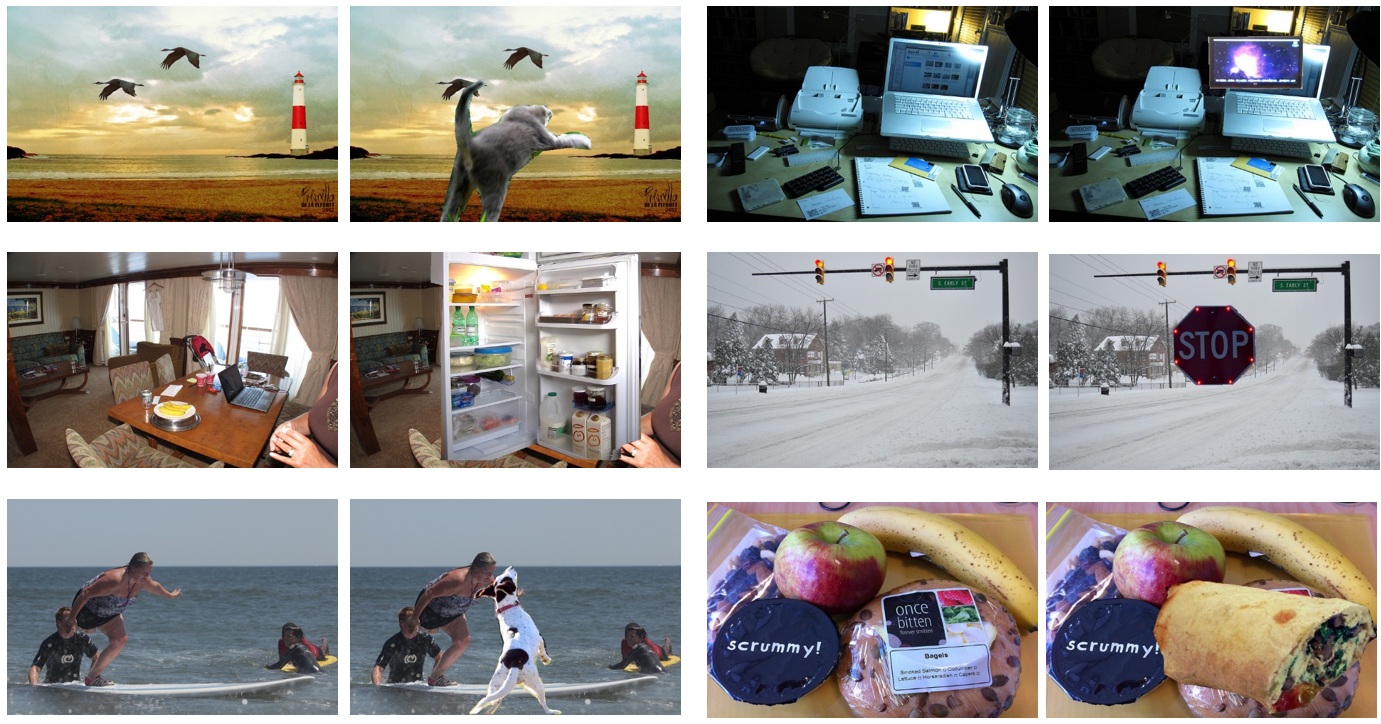}
    \caption{Examples of original (left) vs semantically augmented (right) images.}
    \label{fig:original_vs_augmented_8}
\end{figure*}

\begin{figure*}
    \centering
    \includegraphics[width=0.9\linewidth]{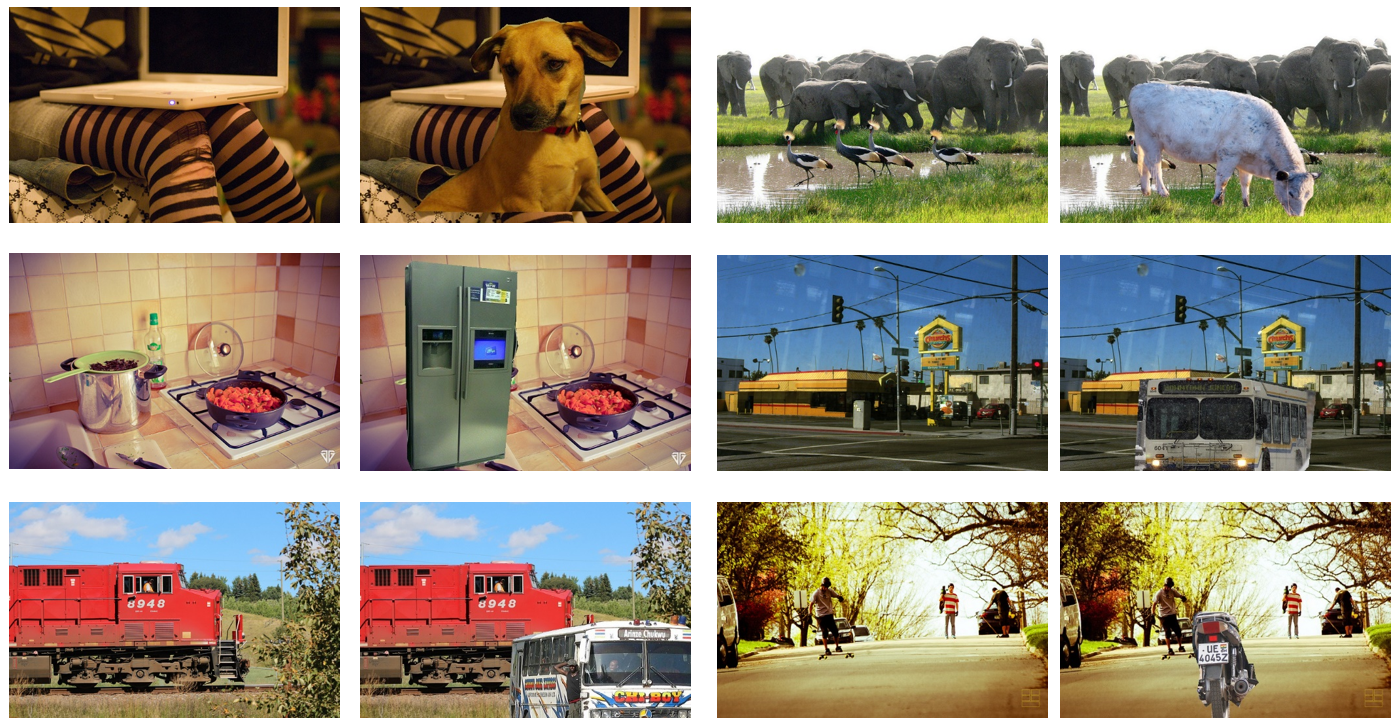}
    \caption{Examples of original (left) vs semantically augmented (right) images.}
    \label{fig:original_vs_augmented_9}
\end{figure*}

\begin{figure*}
    \centering
    \includegraphics[width=0.9\linewidth]{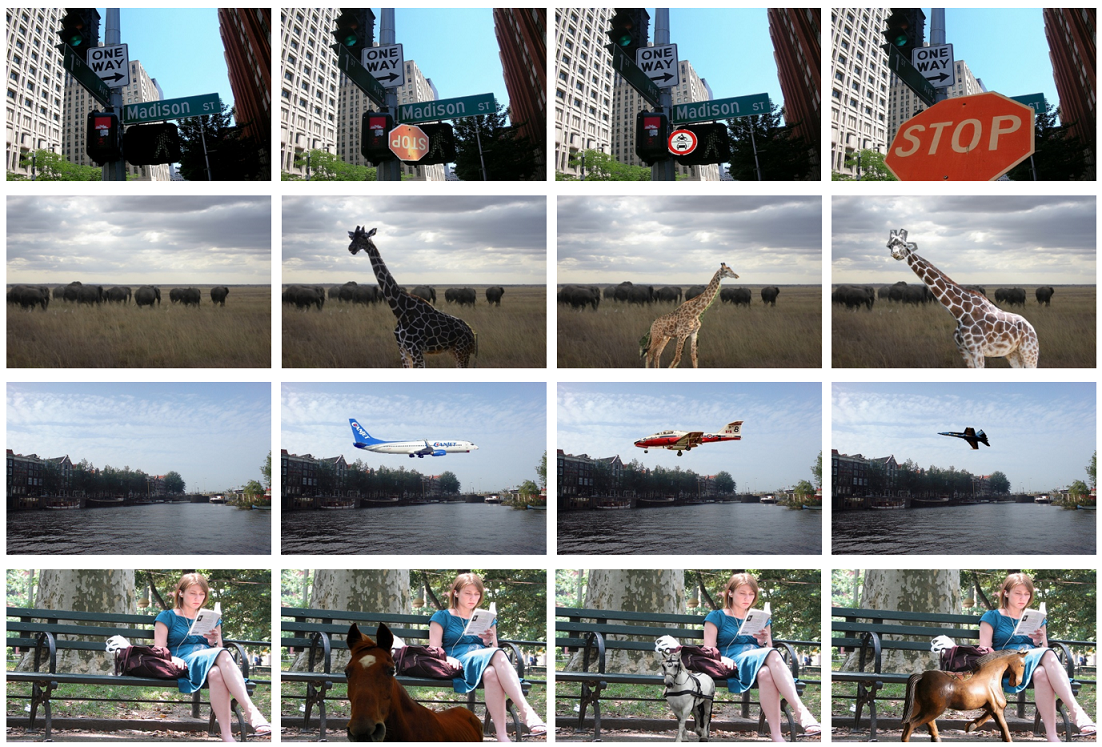}
    \caption{Examples of original (left column) vs semantically augmented images. Different instances of a same object category are being inserted into the host image.}
    \label{fig:same_category_different_instances_1}
\end{figure*}

\begin{figure*}
    \centering
    \includegraphics[width=0.9\linewidth]{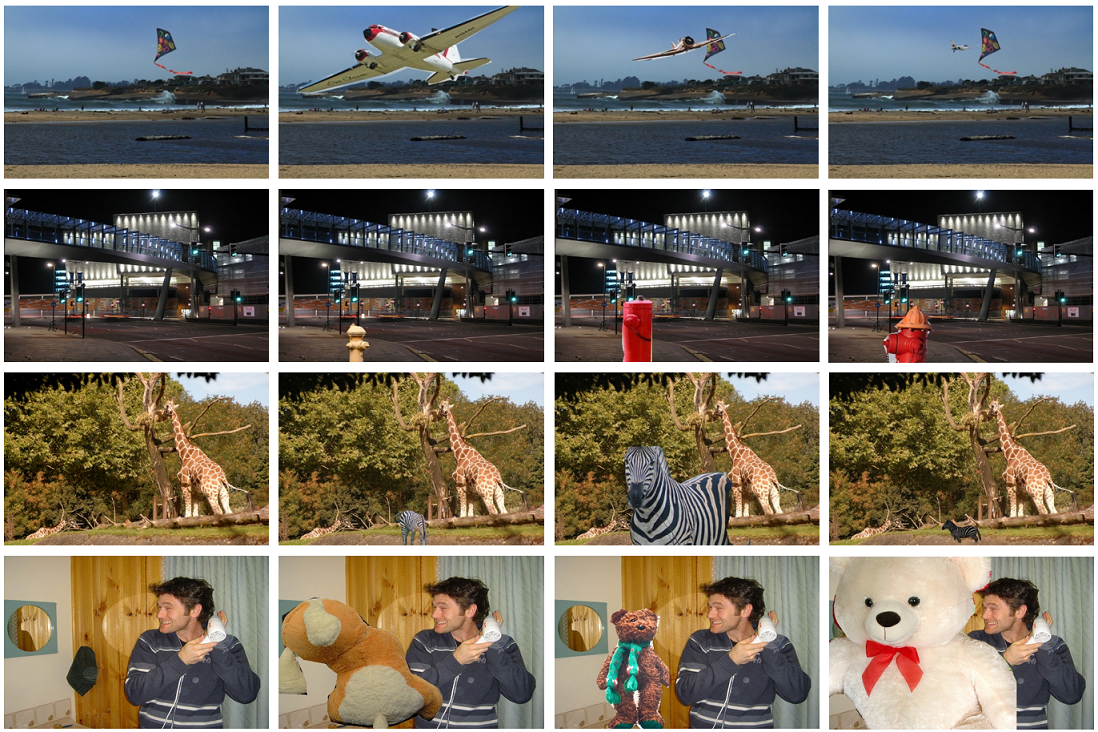}
    \caption{Examples of original (left column) vs semantically augmented images. Different instances of a same object category are being inserted into the host image.}
    \label{fig:same_category_different_instances_2}
\end{figure*}

\begin{figure*}
    \centering
    \includegraphics[width=0.9\linewidth]{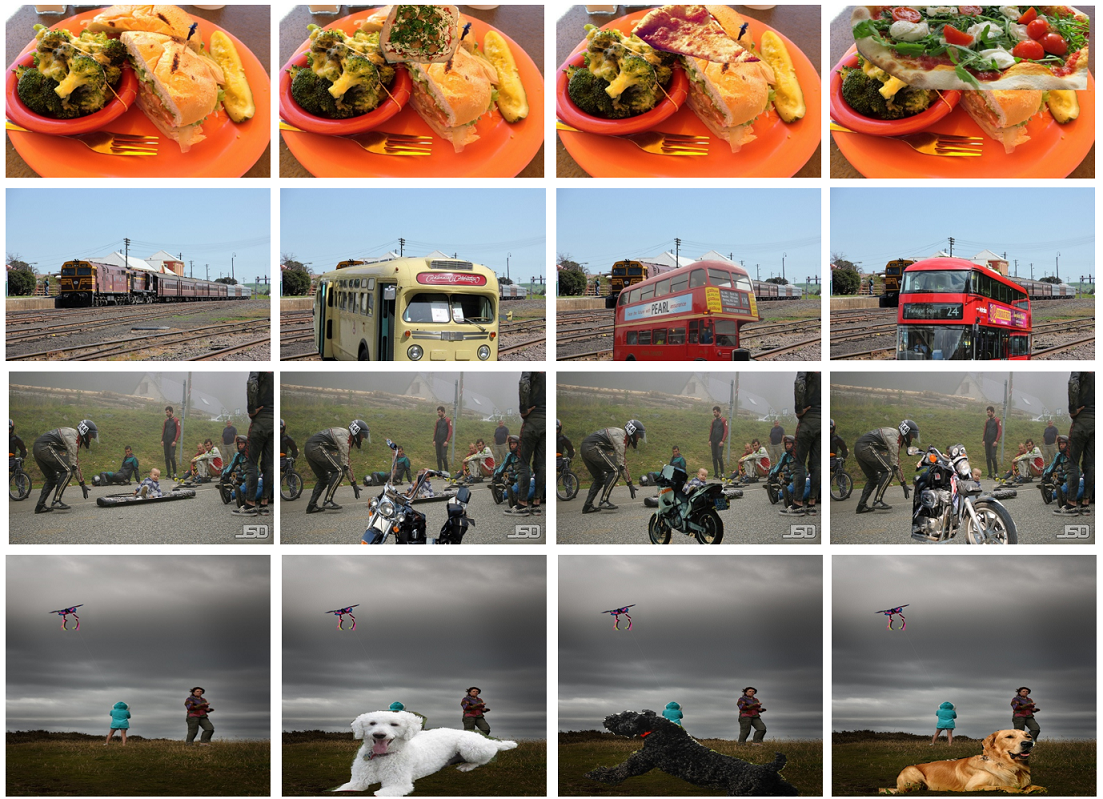}
    \caption{Examples of original (left column) vs semantically augmented images. Different instances of a same object category are being inserted into the host image.}
    \label{fig:same_category_different_instances_3}
\end{figure*}

\begin{figure*}
    \centering
    \includegraphics[width=0.9\linewidth]{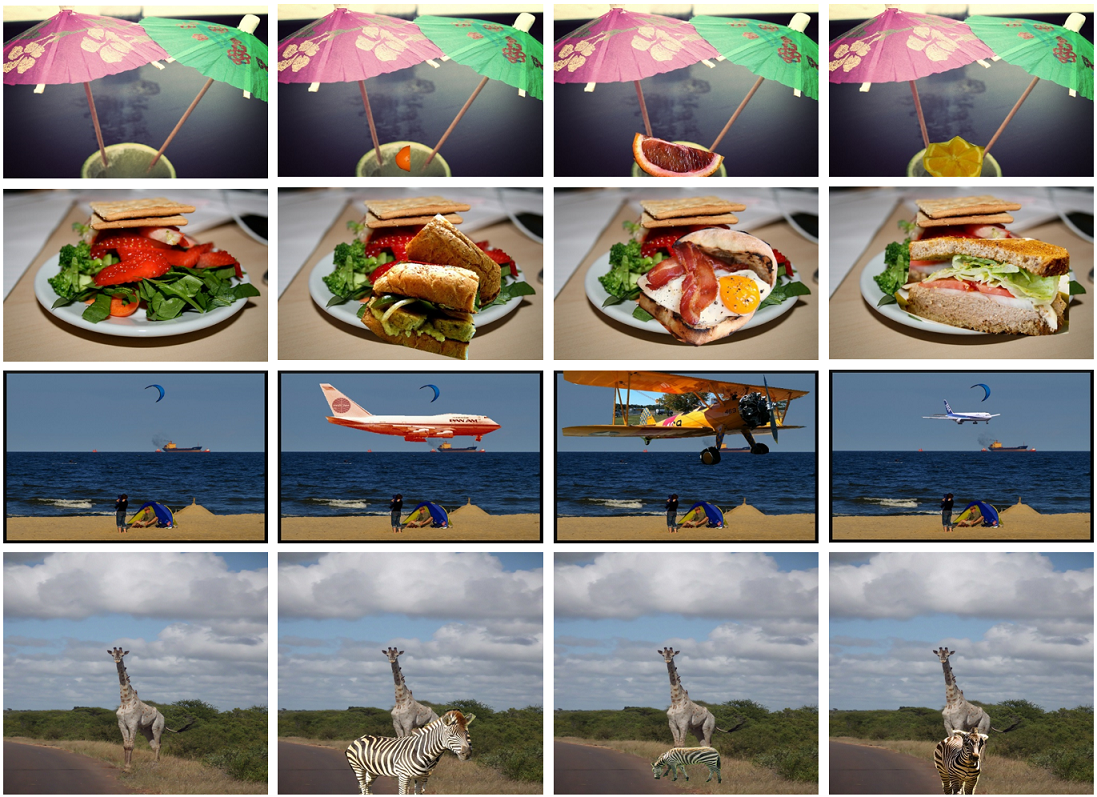}
    \caption{Examples of original (left column) vs semantically augmented images. Different instances of a same object category are being inserted into the host image.}
    \label{fig:same_category_different_instances_4}
\end{figure*}

\begin{figure*}
    \centering
    \includegraphics[width=0.9\linewidth]{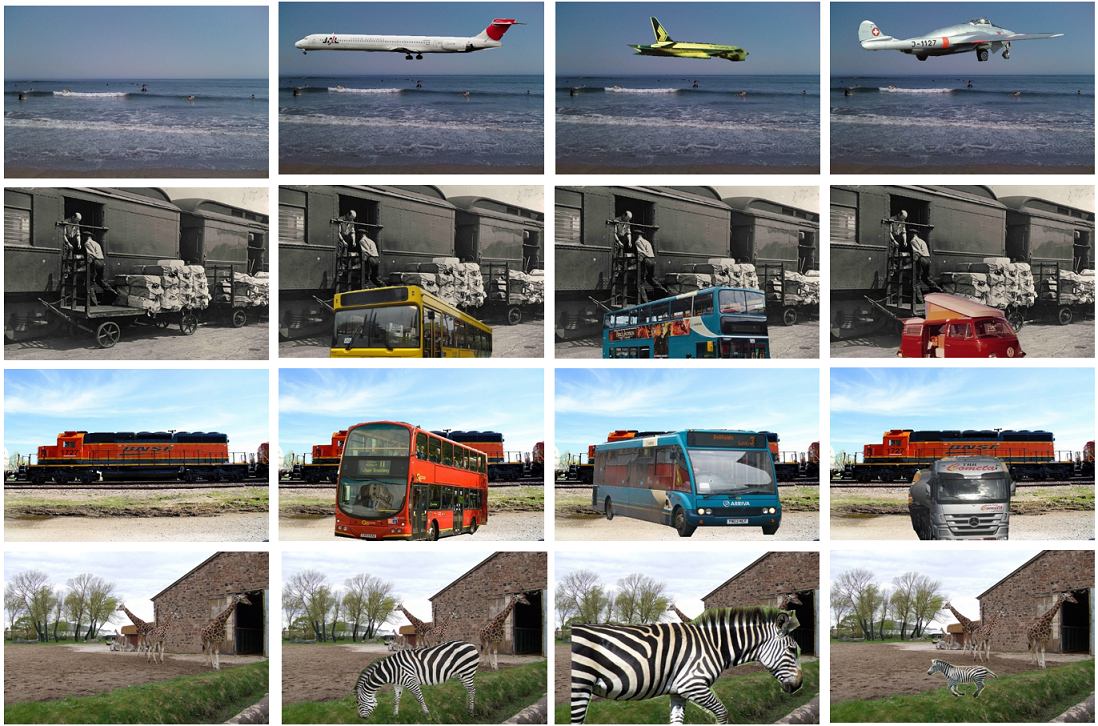}
    \caption{Examples of original (left column) vs semantically augmented images. Different instances of a same object category are being inserted into the host image.}
    \label{fig:same_category_different_instances_5}
\end{figure*}

\begin{figure*}
    \centering
    \includegraphics[width=0.9\linewidth]{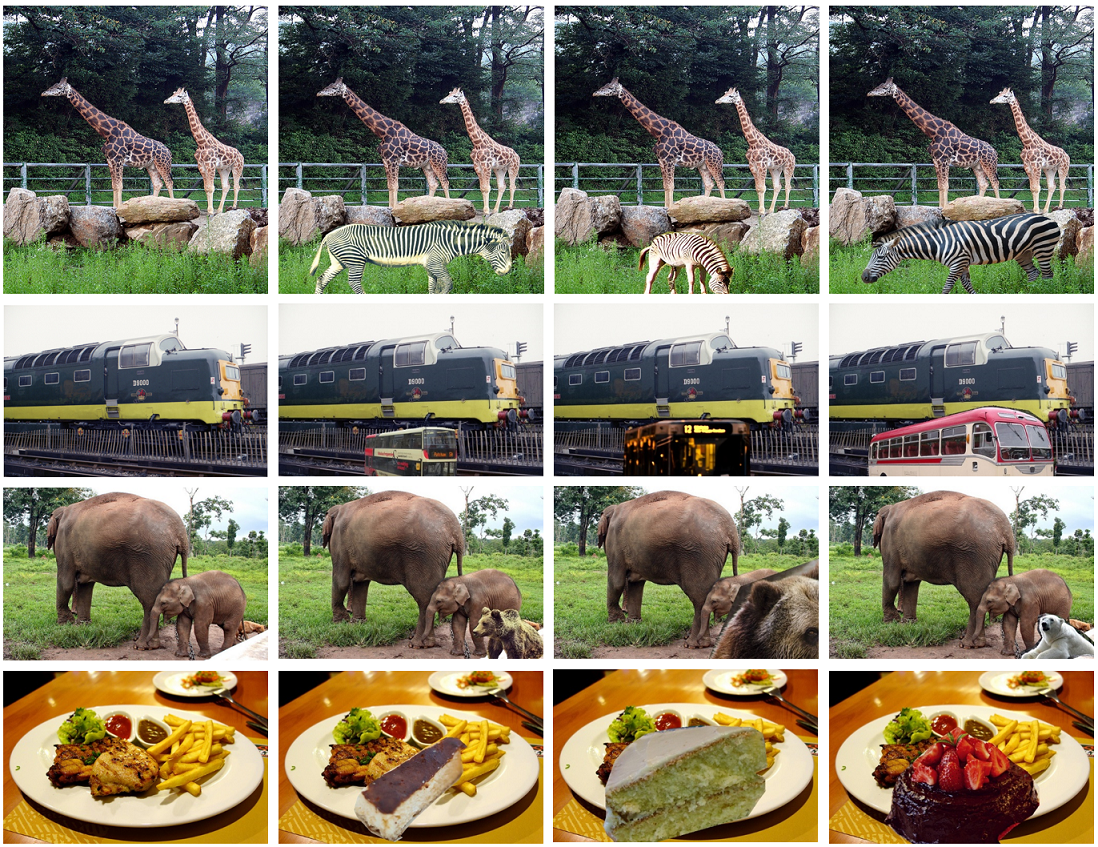}
    \caption{Examples of original (left column) vs semantically augmented images. Different instances of a same object category are being inserted into the host image.}
    \label{fig:same_category_different_instances_6}
\end{figure*}

\begin{figure*}
    \centering
    \includegraphics[width=0.9\linewidth]{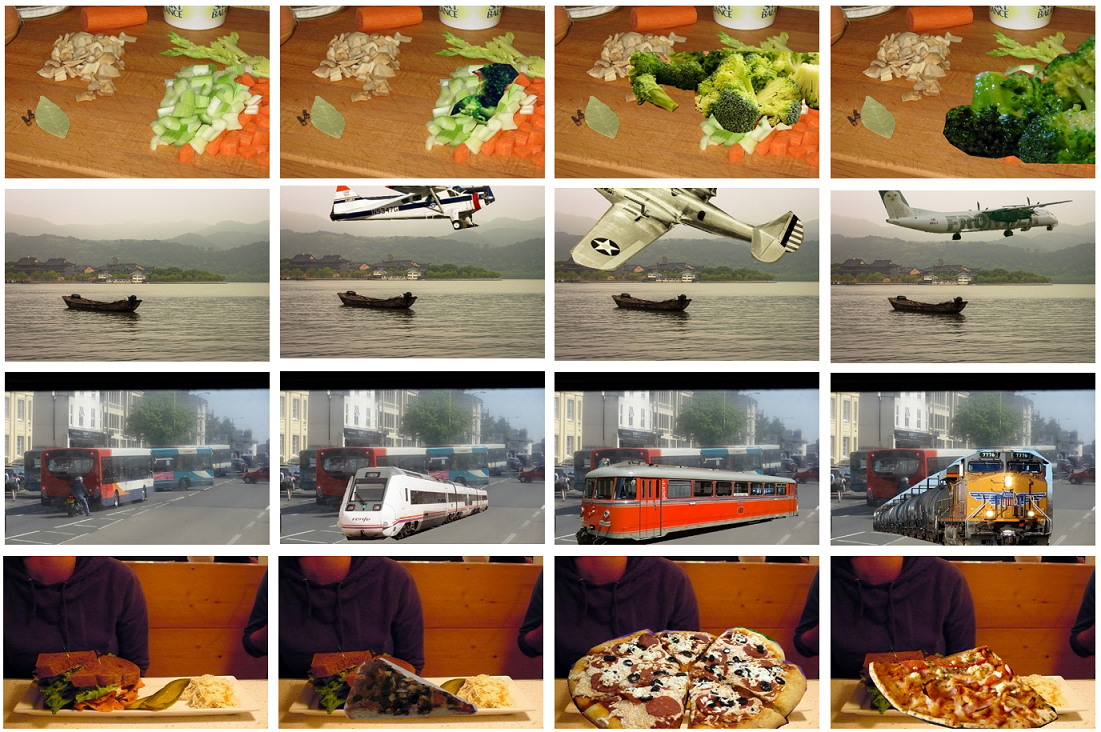}
    \caption{Examples of original (left column) vs semantically augmented images. Different instances of a same object category are being inserted into the host image.}
    \label{fig:same_category_different_instances_7}
\end{figure*}

\begin{figure*}
    \centering
    \includegraphics[width=0.9\linewidth]{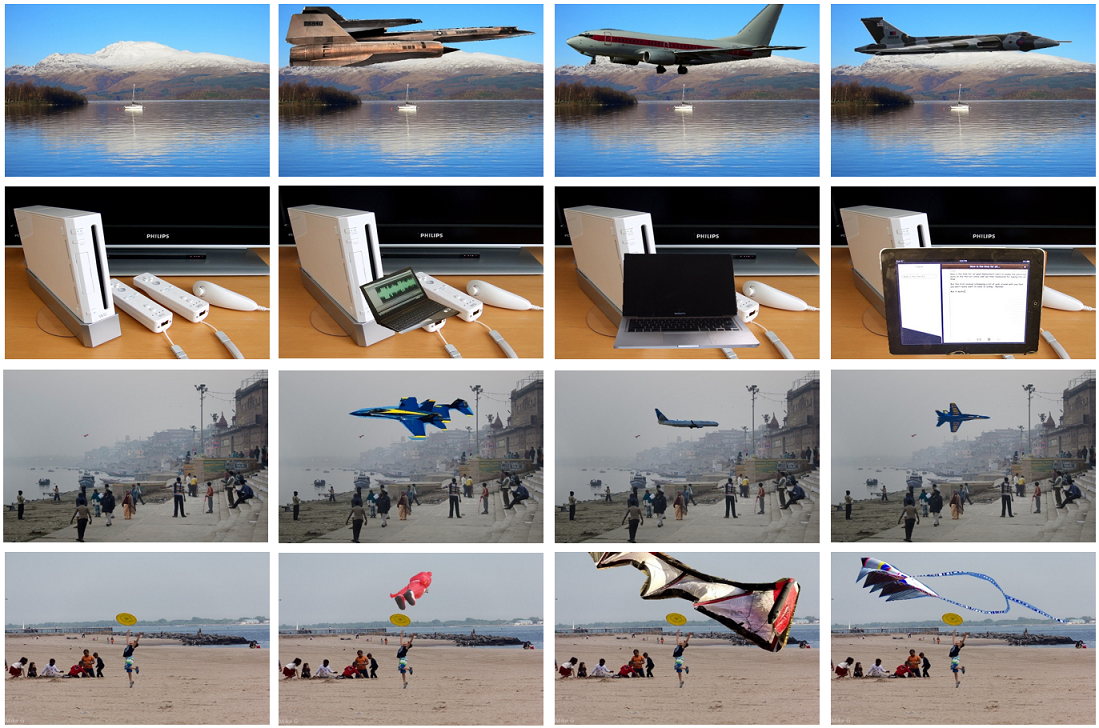}
    \caption{Examples of original (left column) vs semantically augmented images. Different instances of a same object category are being inserted into the host image.}
    \label{fig:same_category_different_instances_8}
\end{figure*}

\begin{figure*}
    \centering
    \includegraphics[width=0.9\linewidth]{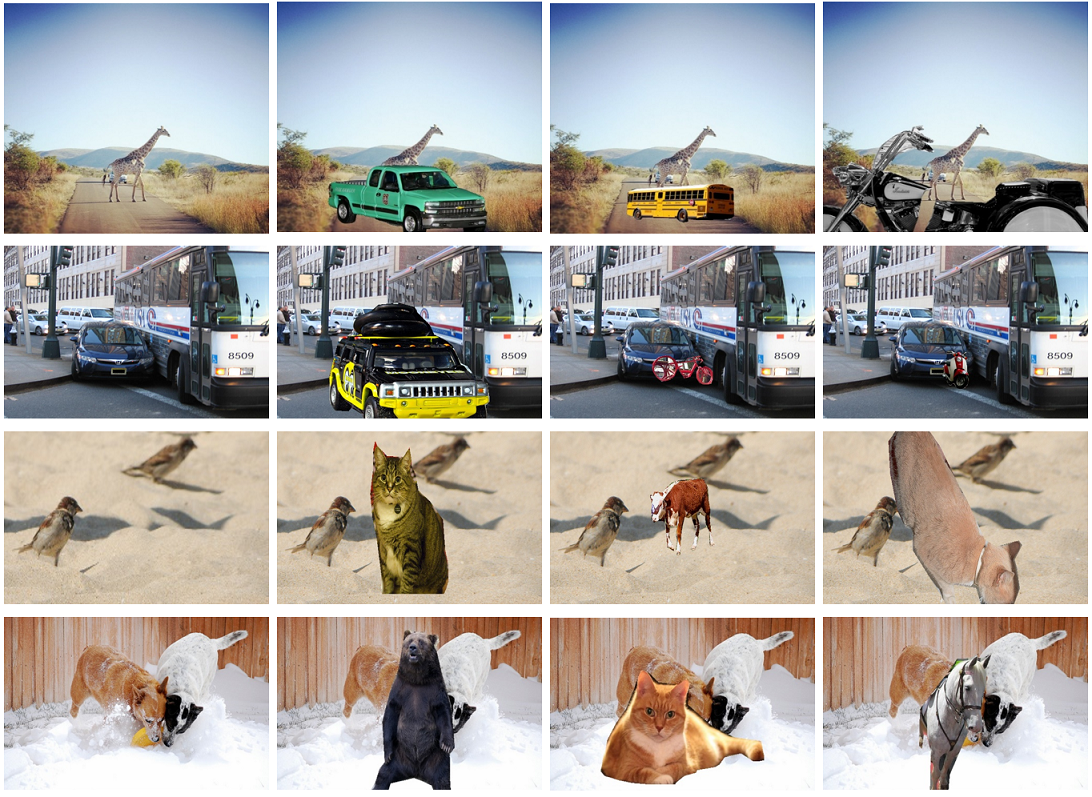}
    \caption{Examples of original (left column) vs semantically augmented images. Different object categories are being inserted into the host image.}
    \label{fig:different_categories_1}
\end{figure*}

\begin{figure*}
    \centering
    \includegraphics[width=0.9\linewidth]{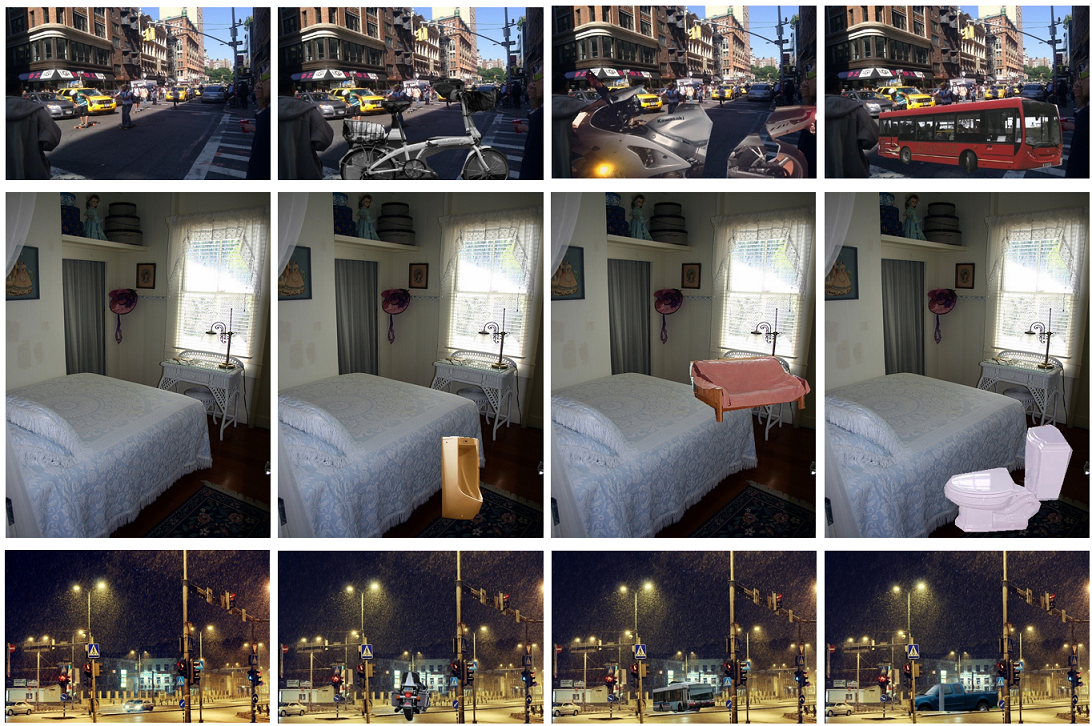}
    \caption{Examples of original (left column) vs semantically augmented images. Different object categories are being inserted into the host image.}
    \label{fig:different_categories_2}
\end{figure*}

\begin{figure*}
    \centering
    \includegraphics[width=0.9\linewidth]{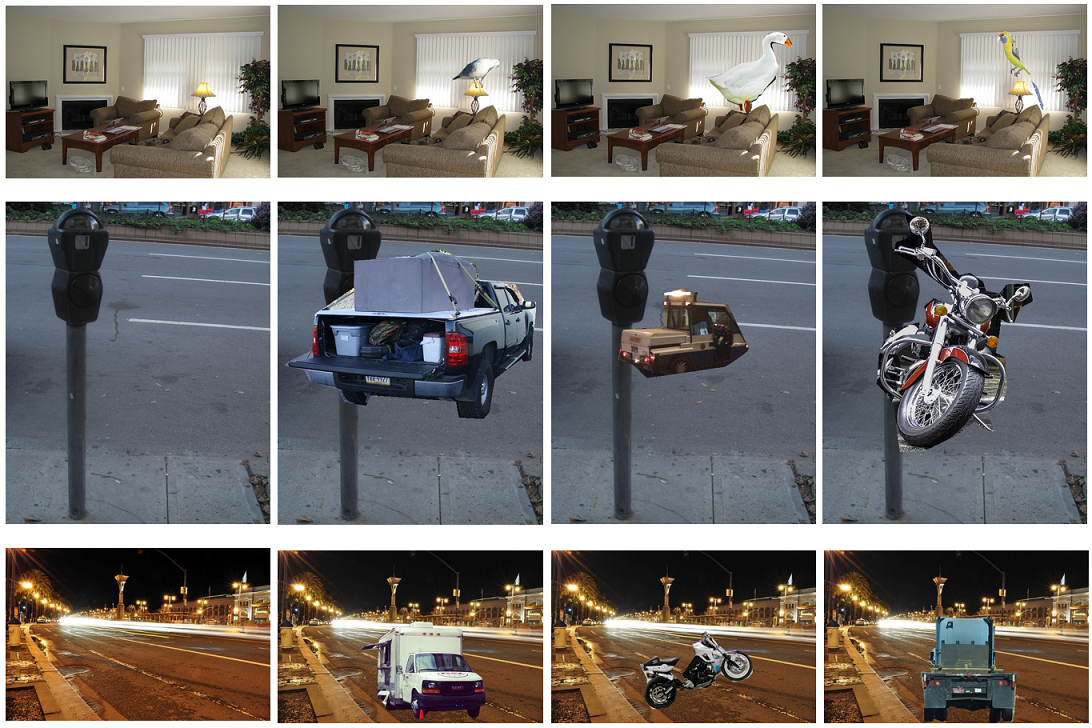}
    \caption{Examples of original (left column) vs semantically augmented images. Different object categories are being inserted into the host image.}
    \label{fig:different_categories_3}
\end{figure*}

\begin{figure*}
    \centering
    \includegraphics[width=0.9\linewidth]{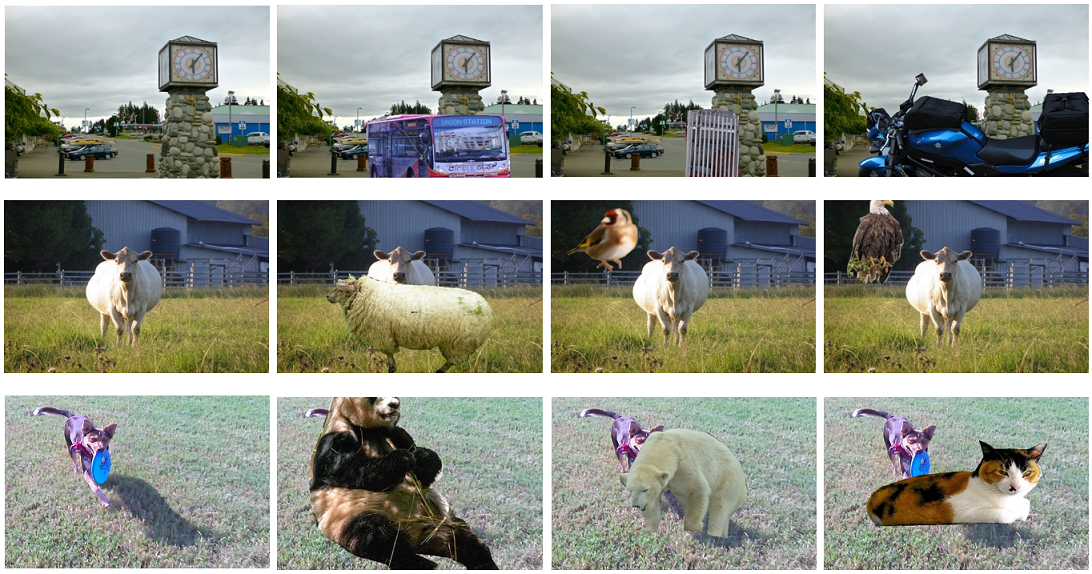}
    \caption{Examples of original (left column) vs semantically augmented images. Different object categories are being inserted into the host image.}
    \label{fig:different_categories_4}
\end{figure*}

\begin{figure*}
    \centering
    \includegraphics[width=0.9\linewidth]{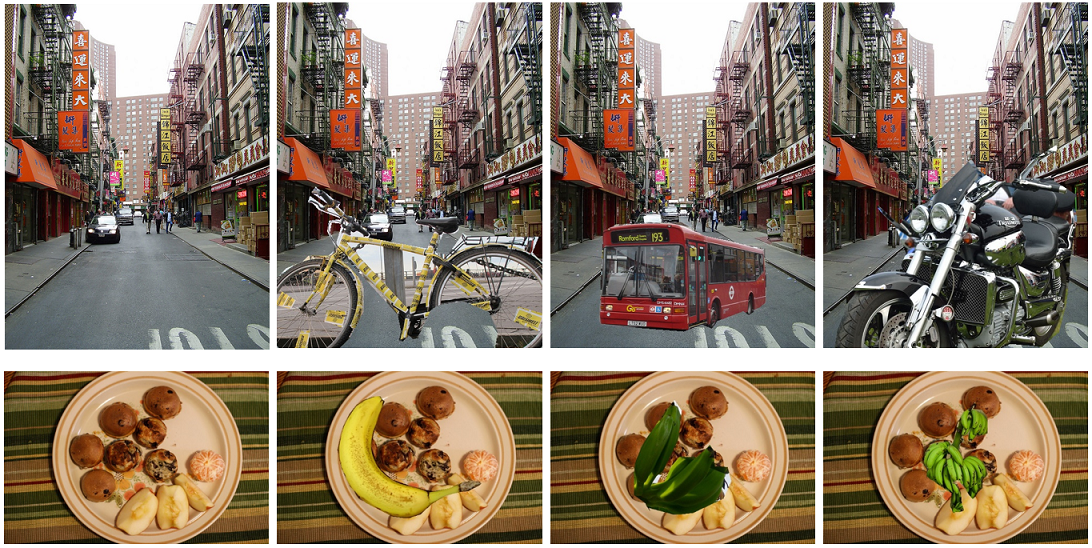}
    \caption{Examples of original (left column) vs semantically augmented images. Different object categories are being inserted into the host image.}
    \label{fig:different_categories_5}
\end{figure*}

Figure \ref{fig:original_vs_augmented_1}-\ref{fig:original_vs_augmented_9} demonstrate additional examples of our semantic augmentation strategy. They show side-by-side comparisons of original versus semantically augmented examples. Moreover, Figure \ref{fig:same_category_different_instances_1}-\ref{fig:same_category_different_instances_8} show examples were different instances from a same object category are selected each time. Figure \ref{fig:different_categories_1}-\ref{fig:different_categories_5} also show the case were different categories are augmented into a same host image.

\end{document}